\documentclass[journal]{IEEEtran}
\usepackage{cite}

\usepackage{dirtytalk}

\ifCLASSINFOpdf
   \usepackage[pdftex]{graphicx}
\else
   \usepackage[dvips]{graphicx}
\fi
\usepackage{hyperref}

\usepackage{amsmath, amssymb}
\usepackage{todonotes}
\usepackage{comment}
\usepackage{amsthm}
\usepackage{enumitem}

\usepackage{algorithm}
\usepackage{algpseudocode}

\makeatletter
\newcommand{\removelatexerror}{\let\@latex@error\@gobble}
\makeatother

\ifCLASSOPTIONcompsoc
  \usepackage[caption=false,font=normalsize,labelfont=sf,textfont=sf]{subfig}
\else
  \usepackage[caption=false,font=footnotesize]{subfig}
\fi
\usepackage{url}

\begin{document}

\onecolumn
{\fontfamily{cmss}
\thispagestyle{empty}
\begin{flushleft}
{\bf\Huge{IEEE Copyright Notice}}
\vspace{20px}

{\Large{\textcopyright 2021 IEEE. Personal use of this material is permitted. Permission from IEEE must be obtained for all other uses, in any current or future media, including reprinting/republishing this material for advertising or promotional purposes, creating new collective works, for resale or redistribution to servers or lists, or reuse of any copyrighted component of this work in other works.}}

\vspace{100px}
{\huge{Accepted to be Published in: IEEE Transactions on Games DOI: 10.1109/TG.2021.3086215.}}

\end{flushleft}}

%
\twocolumn
\title{Evaluating Mixed-Initiative Procedural Level Design Tools using a Triple-Blind Mixed-Method User Study}
%
%
%

\author{Sean~P.~Walton, Alma~A.~M.~Rahat and James Stovold%
\thanks{S. Walton and A.A.M. Rahat are with the Department
of Computer Science, Swansea University, Wales, UK e-mail: s.p.walton@swansea.ac.uk.}
\thanks{J. Stovold is with the Department of Computer Science, British University Vietnam, Vietnam}
}

\maketitle

\begin{abstract}
Results from a triple-blind mixed-method user study into the effectiveness of mixed-initiative tools for the procedural generation of game levels are presented. A tool which generates levels using interactive evolutionary optimisation was designed for this study which (a) is focused on supporting the designer to explore the design space and (b) only requires the designer to interact with it by designing levels. The tool identifies level design patterns in an initial hand-designed map and uses that information to drive an interactive optimisation algorithm. A rigorous user study was designed which compared the experiences of designers using the mixed-initiative tool to designers who were given a tool which provided completely random level suggestions. The designers using the mixed-initiative tool showed an increased engagement in the level design task, reporting that it was effective in inspiring new ideas and design directions. This provides significant evidence that procedural content generation can be used as a powerful tool to support the human design process.
\end{abstract}

\iftrue


%
\setcounter{page}{1}
\IEEEpeerreviewmaketitle

\section{Introduction}
%
%
%
%

\IEEEPARstart{G}{ame} developers are under increasing pressure not only to launch games with hours of unique content, but to continue to add new fresh content post launch~\cite{Roberts2015-sc, Liapis2013-at, Cook2011-vr}. This provides motivation~\cite{Melotti2019-rt, Ruela2018-ah, Baldwin2017-me} to develop tools which can support content generation, which is the aim of procedural content generation (PCG) algorithms~\cite{Van_der_Linden2014-ze}. It is also important to look beyond this commercial motivation and ask how PCG algorithms can support designers in their creative process for the sake of creativity~\cite{Cook2017-gh}.  PCG algorithms have been developed to create a wide variety of content~\cite{Melotti2019-rt}. In addition to supporting designers, PCG algorithms can also benefit the players, resulting in an increased diversity of content~\cite{Melotti2019-rt, Ruela2018-ah, Snodgrass2017-do} and creating a source of curiosity and unpredictability~\cite{Cook2015-qn}. Perhaps the most notable example of this in recent years is Hello Game's title \emph{No Man's Sky}\footnote{\url{https://www.nomanssky.com/}}, a space exploration game in which almost everything is procedurally generated~\cite{Alexandra2016-zd}. There are even examples of using PCG as a game mechanic itself, such as in the game Petalz~\cite{Risi2016-fj} where players breed and share flowers, becoming part of the PCG algorithm itself. 

Despite the clear benefits of PCG algorithms, there are still a number of open challenges in the field. For example, the vast majority of PCG algorithms are highly problem specific, often designed for a single genre of game~\cite{Snodgrass2017-do} or limited to specific geometries~\cite{Von_Rymon_Lipinski2019-tk}. A frequently-cited limitation is the lack of control human designers have when generating content using PCG~\cite{Ruela2018-ah, Von_Rymon_Lipinski2019-tk}. PCG algorithms are often non-intuitive, requiring designers to tweak and adjust tuning parameters which are difficult to relate to their goals. This ultimately limits the control designers have over the generation process~\cite{Van_der_Linden2014-ze} and builds a knowledge barrier~\cite{Baldwin2017-me}.

Despite significant investment into researching new methods for PCG, there is little research on how designers interact with these tools~\cite{Craveirinha2015-hm}. In an attempt to address this gap, Craveirinha and Roque~\cite{Craveirinha2015-hm} undertook a participatory design process involving game designers and researchers to design an interface for a PCG algorithm. In doing so they explored the attitudes of game designers toward PCG tools. They found that many PCG algorithms work by optimising certain metrics which the algorithm designers have identified as being important for player experience. These metrics, and target values for them, are determined \emph{a priori}. Designers do not operate in this way, but instead explore the design space to determine metrics which can then be used to optimise player experience. The findings were summarised with two key observations which will inform our work: \emph{(1) the tool needs an understandable metaphor,} and \emph{(2) exploration is needed before optimisation.}

\subsection{Our Contribution}
\label{sec_pillars}
When investigating the existing work in PCG of levels we found that \textit{most} contributions included \textbf{no user study} of the created artefact~\cite{Alvarez2018-ge, Melotti2019-rt, Ruela2018-ah, Snodgrass2017-do, Baldwin2017-me, Preuss2014-yh, Ashlock2011-xv, McGuinness2011-wu}. In contributions which did include a user study, the studies were performed in an ad-hock fashion with \textbf{no rigorous qualitative analysis} and \textbf{no control group}~\cite{Alvarez2020-xv, Baldwin2017-gc, Liapis2013-at}. \emph{The main contribution of our work is a mixed method triple-blind user study comparing a mixed-initiative PCG tool to a tool which gives the designer random suggestions. Using a reflexive thematic analysis approach~\cite{Braun2019-iv} we find that our mixed-initiative tool does support the creative process, compared to the control group, which adds strength to the findings of the above mentioned studies.} 

The tool designed for this study was rooted in the approach introduced by Baldwin et al.~\cite{Baldwin2017-me}, but placed into the context of the two observations of Craveirinha and Roque~\cite{Craveirinha2015-hm} which we re-framed into two design pillars: 
\begin{enumerate}
    \item The designer must interact with the algorithm by designing content, rather than adjusting parameters.
    \item The designer will be supported to explore the design space.
\end{enumerate}

\section{Background}

\subsection{Search-Based Procedural Content Generation}
\label{sec_search_based_PCG}
There are numerous approaches to PCG~\cite{Togelius2011-xj}. In our work we adopt a search-based approach to PCG as it aligns well with our second design pillar to support exploration. In search-based PCG an algorithm generates a large volume of content and evaluates each item created using a fitness function. There are two key identifying characteristics of a search-based approach: (a) the fitness function allows the comparison and ranking of content, and (b) this ranking is used to inform the generation of new content~\cite{Togelius2011-xj}. Search-based PCG approaches are often implemented using evolutionary algorithms (EAs); optimisation algorithms which aim to minimise a fitness function over several generations. In the context of PCG, an EA will initialise a population of potential designs, rank these according to the quality defined by the fitness function, then create the next generation through stochastic mutation and interbreeding~\cite{Baldwin2017-me}. Search-based approaches have been used to generate a wide range of content including mazes~\cite{Ashlock2011-xv, McGuinness2011-wu}, race tracks~\cite{Loiacono2011-el} and dungeon maps~\cite{Valtchanov2012-kd}. A common aspect of these contributions is that the authors design and specify a fitness function which they argue will result in a good player experience. Our aim is to allow the designer to directly influence the fitness function through design, rather than relying on the fitness function to dictate what is good.

\subsection{Mixed-Initiative Approaches to Content Generation}
As mentioned in the introduction, one of the key challenges of PCG algorithms is that level designers often do not have knowledge of how to control them. This challenge is directly related to our first design pillar, that designers should interact with our system by designing content. Many researchers~\cite{Liapis2013-at, Baldwin2017-me, Ruela2018-ah, Von_Rymon_Lipinski2019-tk} have made contributions towards addressing this challenge. The work by Liapis et al.~\cite{Liapis2013-at} and Baldwin et al.~\cite{Baldwin2017-me} are particularly relevant to our goals and inform our approach.

Liapis et al.~\cite{Liapis2013-at} introduced the Sentient Sketchbook, a tool for supporting designers creating levels for games. As the designer sketches ideas via the tool's interface, real-time feedback is given to the designer based on a number of game play relevant metrics. The tool suggests alternative map designs based on the sketch the designer creates. This is achieved through a genetic search algorithm which attempts to maximise the map's score based on a number of metrics, or a diversity measure. The results of all these searches are presented to the designer. The general feedback from their user study was positive, with users reporting that the tool started pushing them in design directions they did not initially expect. 

Baldwin et al.~\cite{Baldwin2017-me} present a mixed-initiative tool for generating dungeon levels using evolutionary algorithms. Their aim was to allow the designer to control the algorithm using parameters with which they are familiar with, based on what they term \emph{game design patterns}, such as mean corridor length or number of enemies. We suggest a slight change in terminology by referring to these as \emph{level design patterns} hereafter. Game design often refers to the design of mechanics in a game rather than the level geometry, so we feel it is clearer to use the term level design patterns when describing these metrics. In fact, Baldwin et al.~\cite{Baldwin2017-gc} found that the participants in their study found the term game design confusing in this context. Essentially, the level designer specifies targets for the various level design pattern metrics and an evolutionary algorithm attempts to optimise a fitness function based on this. Their results show an impressive ability of control based on these patterns. The tool has been developed over several years since the original paper. More design pattern detection has been implemented~\cite{Baldwin2017-gc} and metrics based on visual aesthetics have been added to the tool~\cite{Alvarez2018-ge}.  Presently, the tool is quite sophisticated, supporting designers to design an entire dungeon rather than single rooms.  Efforts were recently made to model designer preference by training a neural network while a designer interacts with the system~\cite{Alvarez2020-xv}. Although this is an interesting approach which shows promising results, training a neural network to model the preference of a human is a challenging task which requires lots of data. In our work we have decided to take the original approach presented by Baldwin et al.~\cite{Baldwin2017-me} and their findings on the importance of visual aesthetics~\cite{Baldwin2017-gc} and focus on building a tool which only requires the designer to design levels - without tweaking parameters. 

\section{Methodology}

\subsection{Specification and Design of the System/Tool}

\begin{figure}[!t]
\centering
\subfloat[]{\includegraphics[width=2in]{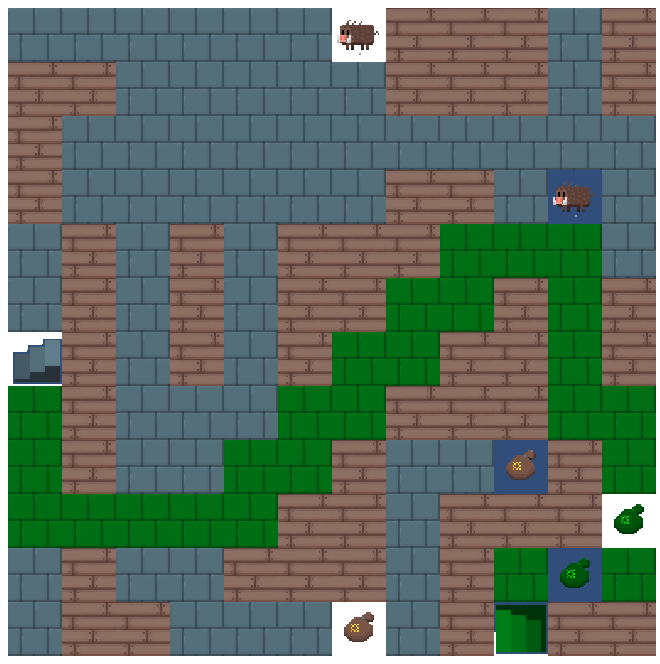}%
\label{fig_tuning_map}}
\hfil
\subfloat[]{\includegraphics[width=2.75in]{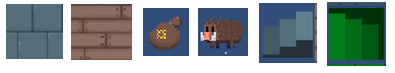}%
\label{fig_key}}
\hfil
\caption{Artwork used to represent map layout and tiles. Assets are distributed by LazerGunStudios without license at~\protect\url{{https://lazergunstudios.itch.io/roguelike-asset-pack}}. In (a), a complete map with a path from entrance to exit is shown, while in (b) we show different adjustable components of the map (Floor, Wall, Treasure, Enemy, Entrance and Exit).}
\end{figure}

A system was designed in the context of two design pillars -- described and justified in Section~\ref{sec_pillars} -- to support a level designer in creating a series of 2D maps/levels for a simple dungeon game. An example of a map is shown in Figure~\ref{fig_tuning_map}. In this study the dungeon maps are made up of 12 by 12 tiles. Each tile has one of six possible values. \emph{Wall}: this is impassable by the player. \emph{Floor}: this is passable by the player. \emph{Treasure}: this is an item which is desirable for the player to reach. \emph{Enemy}: this is a non-player character which can damage the player, something the player wishes to avoid. A single \emph{Entrance} tile where the player enters the level and a single \emph{Exit} representing the players goal. There must be a passable path between the entrance and exit for a level to be valid. The graphical representation of these tiles is shown in Figure~\ref{fig_key}.

\label{sec:PCGApproach}
When surveying the search-based PCG literature we observed two key points:
\begin{enumerate}
    \item Search-based PCG is inherently a multi-objective problem
    \item The majority of researchers tackle this multi-objective problem by combining the results from multiple fitness functions into one scalar value through a weighted sum.
\end{enumerate}
An exception to this is the work by Loiacono et al.~\cite{Loiacono2011-el} who used a multi-objective optimisation algorithm without scalarisation. They found an interesting diversity of solutions along the Pareto fronts, which has the potential to support our second design pillar. Although there are many advanced techniques for multi-objective optimisation and finding the Pareto front~\cite{Zhou2011-fl} we opt for a simple approach which is detailed in Section~\ref{sec_GA}.

Since we wish our designers to interact with our system through designing levels, we turn to the approach by Liapis et al.~\cite{Liapis2013-at} as a starting point. In their approach suggestions are presented to the designer by optimising predetermined fitness functions with the designer's initial design as a starting point. In our approach the level designer will design the first level, the system will then calculate some metrics which describe that level and record those as targets. An evolutionary optimisation algorithm will then randomly initialise a population and try to match the metrics from the user-designed level. Preuss et al.~\cite{Preuss2014-yh} found that restarting their evolutionary algorithm performs as well as advanced approaches to increasing novelty and diversity. Therefore we will restart our algorithm at regular intervals and use this opportunity to allow the level designer to influence the target metrics at run time. This will be achieved by allowing the level designer to edit and select maps produced by the system which are desirable. The system will store the metrics of these \emph{liked} maps and use them in fitness function evaluations.

\subsection{System Overview}

\begin{algorithm}[!t]

\begin{algorithmic}[1]
    \State user designs first level ${{\bf{x}}}_{1}$
    \State store ${{\bf{x}}}_{1}$ in the list of liked maps and the list of levels
    \Repeat
        \State run optimisation algorithm
        \State display a subset of maps from the final generation of the GA 
        \State user may edit maps and tag them as like and/or keep
        \For{each map ${{\bf{x}}}_{i}$}
            \If{${{\bf{x}}}_{i}$ is tagged like or keep}
                \State store ${{\bf{x}}}_{i}$ in the list of liked maps
                \If{user has tagged ${{\bf{x}}}_{i}$ to keep}
                    \State add ${{\bf{x}}}_{i}$ to the list of game levels
                \EndIf
            \EndIf
        \EndFor
    \Until{the list of game levels is full}
\end{algorithmic}
\caption{System Overview}\label{system_code}
\end{algorithm}

\begin{figure}[!t]
\centering
\includegraphics[width=3.5in]{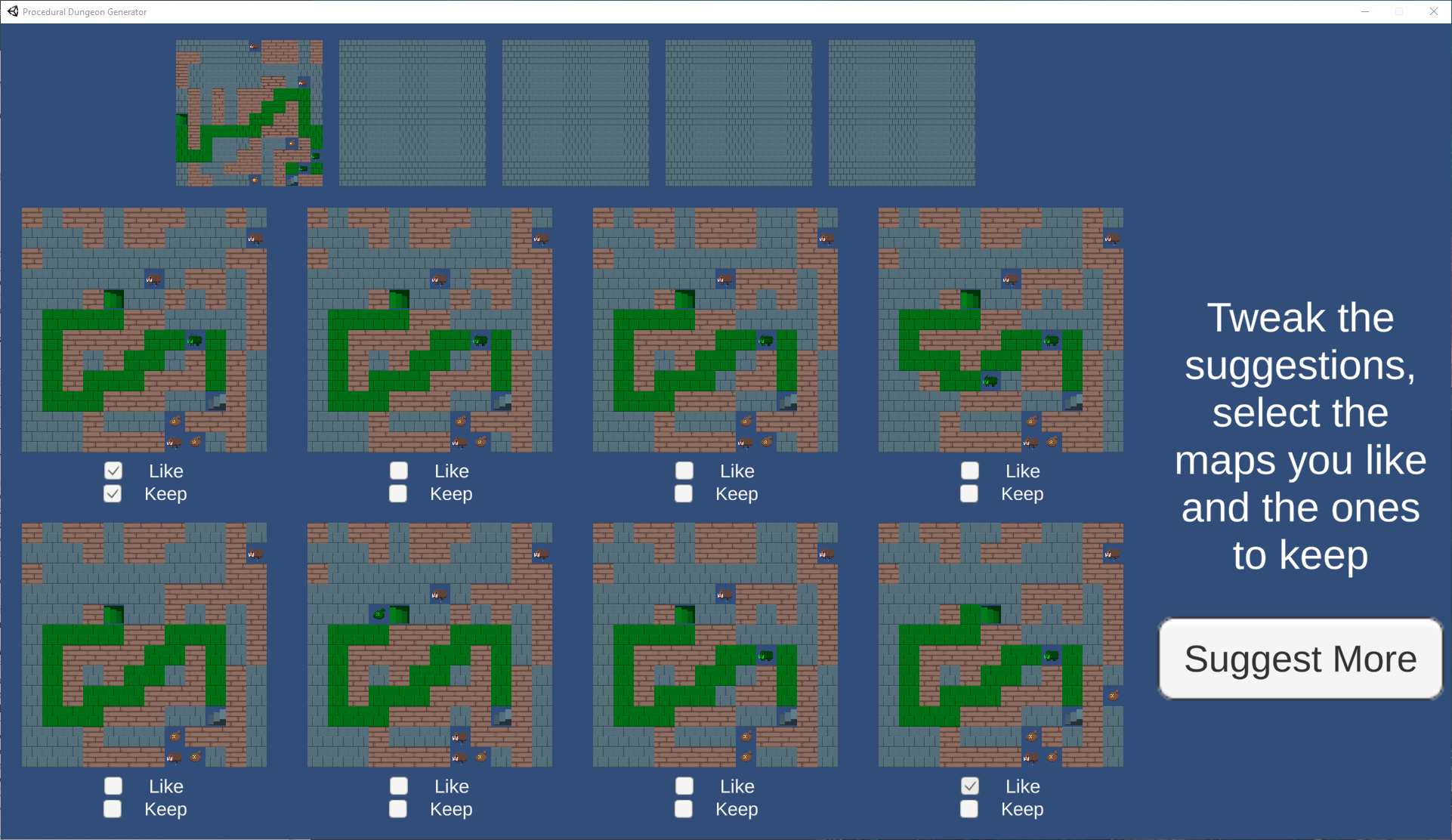}
\caption{Feedback view of the system. At the top row (five smaller windows), the user can see the designs that they have already chosen or created. In middle and bottom rows, we show the generated levels, and provide options for keeping or liking designs. These generated levels can be edited by the user by clicking on their tiles. On the right, the user has the option to request further suggestions.}
\label{fig_present_population}
\end{figure}

Algorithm~\ref{system_code} gives an overview of the final system. The user is initially presented with a blank canvas to design an initial level, once finished the user clicks the submit button. After the optimisation algorithm has finished running the user is presented with the view shown in Figure~\ref{fig_present_population}. The eight maps displayed are a selection from the feasible population of the final generation produced by the optimisation algorithm. Each of the eight maps can be edited by the user; clicking a tile in a map cycles it between all the values possible in turn. The user can additionally tag any number of these maps as \emph{like} or \emph{keep} using the checkboxes below the maps. The \emph{like} tag is used as part of the optimisation process, and the \emph{keep} tag indicates that this map should be included in the final set of levels designed. At the top of the view the levels tagged keep are shown. Once happy with their edits and tags the user clicks "Suggest More" and the tool generates eight more maps replacing those displayed previously. The tool is open source and can be downloaded via GitHub\footnote{\url{https://github.com/seanwalton/mixed-initiative-procedural-dungeon-designer}}.

\subsection{Metrics used to Define the Fitness Functions}
The fitness functions based on level design patterns designed by Baldwin et al.~\cite{Baldwin2017-me} show an impressive ability to control the types of maps generated by their search algorithm. Therefore, we have opted to use these functions along with visual impression metrics which Preuss et al.~\cite{Preuss2014-yh} found to be highly effective. In total there were 31 metrics used to characterise a map design. The metrics are split into two broad categories: level design patterns (\ref{sec:pathlength} to~\ref{sec:treasureMetrics}) and visual impression metrics (\ref{sec:visWallSym} to~\ref{sec:exactSym}). We use the notation that $M_{i}({\bf{x}}_{k})$ is metric $i$ calculated for the map ${\bf{x}}_{k}$. It is worth emphasising that we are not optimising these metrics directly, but using them to construct the fitness function which will drive the optimisation process.
\subsubsection{Path Length}
\label{sec:pathlength}
$M_{1}$ is the path length, $P(Entrance,Exit)$, measured in number of tiles, divided by the total number of tiles in the map, $N_{total}$.

\subsubsection{Global Wall to Passable Tile Ratio}
$M_{2}$ is the ratio of walls to non--wall tiles in the map.

\subsubsection{Corridor Metrics}
\label{sec:corridorMetrics}
Corridors are defined as horizontal or vertical series of passable tiles enclosed by impassible tiles on either side~\cite{Baldwin2017-me}. In our implementation corridors of length one are counted. The metrics $M_{3}$ to $M_{6}$ are the number of corridors followed by the maximum, minimum and mean corridor lengths.

\subsubsection{Chamber Metrics}
A chamber is defined as a continuous block of passable tiles which are wider than a corridor. A less rigid definition is followed than the one outlined by Baldwin et al.~\cite{Baldwin2017-me}. In their work these metrics are used to generate dungeons using user inputs such as chamber size, therefore they have to consider what a user might expect a chamber to look like. In our work these metrics are only used to compare the structure of two maps, we do not want to assume a minimum chamber size. Chambers are identified following corridor identification.  Once chambers are identified two qualities for each chamber is calculated, the area $k^{A}_{i}$ and squareness $k^{S}_{i}$ given by:
\begin{equation}
    k^{A}_{i} = k_{i}^{h}k_{i}^{w}
\end{equation}
\begin{equation}
    k^{S}_{i} = \frac{k^{A}_{i}}{{\min(k_{i}^{h}, k_{i}^{w})}^2}   
\end{equation}
Where $k_{i}^{h}$ and $k_{i}^{w}$ are the height and width of chamber $i$. This then leads to 7 metrics ($M_{7}$--$M_{13}$) for chambers. The total number of chambers, the maximum, minimum and mean chamber areas and maximum, minimum and mean chamber squareness.

\subsubsection{Dead Floor Tiles}
A dead tile is defined as a passable tile which has not been identified as a chamber or corridor. These often appear as tiles which connect multiple corridors or chambers. The metric, $M_{14}$ is simply the number of these tiles divided by the total number of tiles.

\subsubsection{Entrance Metrics}
Two metrics are defined for the number of treasure and enemy tiles around the entrance~\cite{Baldwin2017-me}. $M_{15}$ is the minimum area around the entrance tile which does not contain an enemy tile, and $M_{16}$ is the minimum area around the entrance tile which does not contain a treasure tile.

\subsubsection{Enemy and Treasure Metrics}
\label{sec:treasureMetrics}
$M_{17}$ and $M_{18}$ are simply the fraction of enemy and treasure tiles respectively. In addition a safety measure, defined in~\cite{Baldwin2017-me}, is calculated for each treasure and $M_{19}$ and $M_{20}$ are the mean and standard deviation of this.

\subsubsection{Visual Symmetry of Wall Tiles}
\label{sec:visWallSym}
Preuss et al.~\cite{Preuss2014-yh} introduced a number of visual symmetry metrics which we have adapted for use here. Two lines of symmetry are defined along the centre of the map horizontally and vertically. The number of a specific type of tile is counted either side of these lines then used to calculate ratios. For example, $$N_{wall}^{top}$$ is the number of wall tiles in the top half of the map and $$N_{wall}^{left}$$ is the number of wall tiles in the left half of the map. A total of 8 metrics are defined based on these ratios, for example the left to right wall tile ratio is:
\begin{equation}
\label{ref:equ_visSym1}
    M_{21}({\bf{x}}_{k}) = \frac{\left|N_{wall}^{left}-N_{wall}^{right}\right|}{N_{wall}}
\end{equation}
There is also a top to bottom wall ratio $M_{22}$, left to right and top to bottom enemy and treasure ratios ($M_{23}$--$M_{26}$), and treasure to enemy ratios defined as:
\begin{equation}
    M_{27}({\bf{x}}_{k}) = \frac{\left|N_{treasure}^{left}-N_{enemy}^{right}\right|}{N_{treasure}+N_{enemy}}
\end{equation}
\begin{equation}
\label{ref:equ_visSym2}
    M_{28}({\bf{x}}_{k}) = \frac{\left|N_{treasure}^{top}-N_{enemy}^{bottom}\right|}{N_{treasure}+N_{enemy}}
\end{equation}
For equations~\ref{ref:equ_visSym1} to~\ref{ref:equ_visSym2} if the denominator would be zero the metric is given a value of zero.

\subsubsection{Exact Symmetry Metrics}
\label{sec:exactSym}
As well as the visual symmetries we also introduce and define 3 metrics which give a measure of exact reflection over the symmetry lines used for ($M_{21}$--$M_{28}$). In addition a measure of rotational symmetry is considered by comparing the map against its transpose. These metrics ($M_{29}$--$M_{31}$) are calculated by simply counting the number of tiles that exactly match their reflected counterpart across the various symmetry lines (or to the tile at its transposed location) and express them as a fraction of the total number of tiles.





\subsection{Genetic Algorithm} \label{sec_GA}
Following the approach of Baldwin et al.~\cite{Baldwin2017-me} we use a feasible--infeasible two-population (FI-2Pop) genetic algorithm (GA)~\cite{Kimbrough2008-mj} as our evolutionary optimisation algorithm. FI-2Pop is the same as a standard GA but splits the population into feasible and infeasible sub-populations. Maps are considered feasible if a valid path from the entrance to exit exists, and are automatically entered into the correct sub-population once created. In our system the only difference between evaluating fitness in these two populations is that $M_{1}$, the path length, is not considered for the infeasible maps. A tournament selection approach is used to select individuals (from the same sub-population) to reproduce and a number of elite individuals survive from one generation to the next. 
\subsubsection{Fitness Functions and Ranking Procedure} \label{sec:domination}
The fitnesses of an individual map are defined as
\begin{equation}
    f_{i}({\bf{x}}_{k}) =  \min_{t = 1}^T\left| M_{i}({\bf{x}}_{k}) - M_{i}({\bf{x}}_{t}) \right|
\end{equation}
where $i \in [1,31]$, ${\bf{x}}_{t}$ is one of the $T$ levels a user has liked or stored. This is a form of goal programming approach, where our aim is to generate maps that have metrics similar to those provided by the designer. Thus, lower fitness values are desirable.

Map $\bf{x}_j$ is said to dominate map $\bf{x}_{k}$, denoted as $\bf{x}_j \prec \bf{x}_k$, iff $f_{i}(\bf{x}_{j}) \leq f_{i}(\bf{x}_{k})$ for all $i \in [1, 31]$, and there is at least one fitness function $f_l$ for which $f_{l}(\bf{x}_{j}) < f_{l}(\bf{x}_{k})$ \cite{deb:multi}. The Pareto set of mutually non-dominated solutions is defined as:
\begin{align}
\label{eq:ps}
\mathcal{P} = \{\bf{x} ~|~ \bf{x}' \nprec \bf{x}, \forall \bf{x}, \bf{x}' \in \mathcal{X} \wedge \bf{x} \neq \bf{x}'\},
\end{align} 
where  $\mathcal{X}$ is the feasible decision space. Typically, it is impossible to exactly locate the Pareto set, so a representative approximation, $\mathcal{P}^*$, is often sufficient.

Optimising more than three objectives simultaneously is referred to as a many-objective optimisation problem~\cite{li2018evolutionary}. As the number of objectives increase so does the probability of finding solutions which improve at least one of them, compared to existing solutions, leading to almost all solutions in the search space becoming Pareto optimal~\cite{coello:evolutionary}. Thus locating a representative approximation of the Pareto set is extremely challenging~\cite{walker:tevc}. An \textit{ad hoc} approach to identify a smaller subset of $\mathcal{P}^*$ is often required. Osyczka \textit{et al.} proposed to prune $\mathcal{P}^*$ such that only one solution within a predefined interval is retained and others in close proximity are deleted \cite{osyczka2001evolutionary}. As identifying a sensible interval is not straightforward in a high-dimensional objective space, such an approach may still result in a large number of solutions in the $\mathcal{P}^*$. 

In this paper, we, therefore, take a more aggressive approach in order to keep candidates of interest in our population. We aim to only retain maps that are \textbf{more similar} to the ones selected by the user: this is a \textit{majority voting} approach, popular in decision making with ensemble of systems~\cite{polikar2006ensemble}. We deliberately ignore the magnitude of the differences in various objectives for two reasons. Firstly, the standard euclidean distance in high-dimensions loses its efficacy in objectively identifying how different two maps are \cite{aggarwal2001surprising}. Secondly, it is not obvious if there is a natural preference between objectives. As such our \textit{ad hoc} measure for ranking and retaining solutions in the population is: given $\bf{x}_j, \bf{x}_k \in \mathcal{X}$, if more $f_{i}({\bf{x}}_{j}) \leq f_{i}({\bf{x}}_{k})$ than $f_{i}({\bf{x}}_{k}) \leq f_{i}({\bf{x}}_{j})$ for $i \in [1,31]$, then we prefer $\bf{x}_j$ over $\bf{x}_k$.

\paragraph{Majority Voting and Pareto Optimality}

For majority voting to be effective in ranking and retaining solutions of interest, it must have the following characteristic:
\begin{description}
\item[$C$.] The dominance relationship must be preserved, that is if for two arbitrary solutions $\bf{x}$ and $\bf{x'}$, $\bf{x} \prec \bf{x'}$, then majority voting must rank $\bf{x}$ better than $\bf{x'}$.
\end{description}

\noindent
\begin{proof}
Comparing two solutions $\bf{x}$ and $\bf{x}'$ with $K$ objectives, we can evaluate sets of the following relationships: $\mathcal{L} = \{i\in[1, K] \subseteq \mathbb{N}_1 | f_i(x) < f_i(x')\}, \mathcal{E} = \{j\in[1, K] \subseteq \mathbb{N}_1 | f_j(x) = f_j(x')\}$ and $\mathcal{G} = \{k\in[1, K] \subseteq \mathbb{N}_1 | f_k(x) > f_k(x')\} $, where $i\neq j\neq k$ and $\mathbb{N}_1$ is the set of positive integers.

According to majority voting, we can conclude that $\bf{x}$ is better than $\bf{x'}$ iff:
\begin{align}
\label{eq:psp}
|\mathcal{L}| + |\mathcal{E}| > |\mathcal{G}|.
\end{align}

By definition of Pareto optimality (see \eqref{eq:ps}), $|\mathcal{G}| = 0$ when $\bf{x} \prec \bf{x'}$. Hence, for any number of objectives greater than $0$, \eqref{eq:psp} must be true, and thus $\bf{x}$ will be ranked higher than $\bf{x'}$. Thus, $C$ holds.
\end{proof}

\paragraph{Implications on the Search Population}

We start our search with an arbitrary population of fixed size. Each individual in the population is ranked using the majority voting approach. Since $C$ holds, the best ranked individual is the best approximation of a Pareto optimal solution in the population, and there may be mutually non-dominated solutions in the population based on the balance reached between the sizes of the sets $\mathcal{L}$, $\mathcal{E}$ and $\mathcal{G}$.

When we update the population, we only allow a child $\bf{x}$ to replace a solution $\bf{x}'$ in the population if and only if $\bf{x}$ is better ranked. Clearly, this means one of following two possibilities is met for replacing an individual in the population:
\begin{enumerate}[label=(\roman*)]
    \item $\bf{x \prec \bf{x}'}$, i.e. individual is dominated by the child. 
    \item $\bf{x \nprec \bf{x}'}$, i.e. individual and child are mutually non-dominated, but child satisfies \eqref{eq:psp}.
\end{enumerate}
This way we are creating evolutionary pressure to eradicate dominated individuals from existing population. Thus we are highly likely to retain only mutually non-dominated solutions in the final population, and generate a fixed sized approximation of $\mathcal{P}$.

\subsubsection{Crossover and Mutation} 
\paragraph{Crossover} 
Two parent maps ${\bf{x}}_{i}$ and ${\bf{x}}_{j}$ are crossed to create a child map ${\bf{x}}_{c}$ by first randomly picking an entrance and exit from the parents such that ${\bf{x}}_{c}$ has exactly one entrance and one exit. The remaining tiles in ${\bf{x}}_{c}$ are randomly selected from ${\bf{x}}_{i}$ and ${\bf{x}}_{j}$ with equal probability.

\paragraph{Mutation}
To mutate ${\bf{x}}_{c}$, a random tile is selected and then swapped with a random adjacent tile to create the mutated map.

\subsubsection{Selecting Tuning Parameters}
The performance of GAs is highly problem dependent~\cite{Wolpert1997-mp}. It is therefore crucial to carry out a parameter sensitivity study for each new application to maximise performance.~\cite{Walton2019-ao}. The map shown in Figure~\ref{fig_tuning_map} was selected for use in the parameter study since it has a balance of corridors and chambers. For brevity we do not present the detailed results of our studies, but explain our process and present the final parameters used. For each combination of parameters we performed $30$ tests with different random seeds for the random number generator. We then compared mean performance to select the final set of parameters. It is challenging to define good performance in the context of our aims which are primarily human focused. As an approximate measure we compared the sum of fitnesses of the best level generated for different combinations of parameters. In all tests the number of objective function evaluations was kept constant at $10,000$. This was a decision made based on the time taken for an optimisation run to complete on the machines used in the user study, with the aim of limiting the participation time of the user study to $30$ minutes. The best performing set of parameters and methods were found to be: Mutation Rate: $0.5$, Tournament Size: $2$, Number of Elite: $1$, Population Size: $20$ and Number of Generations: $500$. Examples showing how the GA performs when driven without human input are provided in the supplementary materials for this paper.

\section{User Study}

\subsection{Methodology}
Yannakakis et al.~\cite{Yannakakis2014-rt} introduce an assessment methodology for mixed-initiative systems. They recommend evaluating how often the computational creations are used by the designer, and whether or not those creations changed the thinking process of the designer; our user study was designed to evaluate these aspects. Ethical approval for the study was obtained from the Swansea University College of Science ethics committee\footnote{SU-Ethics-Staff-100220/214}. Our original plan was to perform the study in lab conditions, however due to the COVID-19 pandemic we had to change our methodology and carry out the study on-line. Four participants did complete the study in lab conditions prior to the UK lock-down, every effort was made to ensure parity between the lab and on-line experiments. Participants were recruited through social media and the research team's professional networks. The only requirements for taking part in the study were that you had to be aged 18 or over and have access to an internet-connected computer running Windows or Linux. Participants were each given an information sheet which explained that we were investigating approaches people take when designing levels for video games, with the aim to better understand this process to enable us to make level design tools. They were then asked to \say{create 5 levels for a simple dungeon game using a computer assisted tool.} A set of instructions for using the tool and what constituted a valid level were provided along with the tool itself.

Some slight modifications to the tool were made for the user study. Alongside the suggestions from the system, participants were given a blank canvas where they could design a new level from scratch. Before starting the process the tool asked the participant to enter a unique ID. Based on this ID there was a 50\% chance the tool used the GA designed in this paper to generate suggested maps. In all other cases maps were randomly generated with no optimisation at all. This was done using a triple-blind approach, neither the participant or researchers knew which algorithm had been selected until after the data was analysed. The result is that we have two groups of participants to compare, the GA group and the control group (who were given random suggestions). Once the participant completed the game design task they were asked to upload log files which contained quantitative results and answer a series of free response questions.

\paragraph{Quantitative Measures}
Each participant submitted a log file which contained the following quantitative measures:
\begin{itemize}
    \item Which participant group they belong to (GA or control)
    \item The number of maps the participant marked as like or keep at each iteration.
    \item The number of times the participant created a map from scratch using the blank editor.
    \item How much a participant tweaked a suggested design if they decided to keep or like it.
\end{itemize}
Participants were also required to submit a screenshot of the final screen of the tool, which includes the 5 levels they created, these are included in the supplementary materials for this paper.

\paragraph{Qualitative Questions}
Each participant was then asked 4 questions with a free text response. The questions were:
\begin{enumerate}
    \item Describe the process you took to design a new level.
    \item Was designing 5 levels challenging, or could you have easily designed many more? Explain your answer.
    \item Did the tool affect the way you designed your levels? Explain your answer.
    \item How would you describe the tool to someone else? 
\end{enumerate}
To analyse the responses an inductive coding approach was adopted. Codes were created by reading through all responses, to all questions, independently by each member of the research team. These codes were combined into a final set of codes for each question, which were used for the final coding which was performed by SW. This analysis was all carried out before participant responses were linked to their group, making our study triple-blind.

\subsection{Materials}
\label{mat}
A total of 24 participants took part in the study. Of those 17 (71\%) were male, 6 (25\%) female and one (4\%) did not disclose their gender. The mean age of participants was 25.2 years (SD = 7.81, range = 18 to 48). Participants were asked two questions relating to the frequency with which they play video games and their experience with designing levels. The majority (83\%) of participants reported that they play games frequently, more than once a month. Participants rated their level design experience on a Likert scale from 1 \emph{No Experience} to 5 \emph{Level Design is my primary profession}. The mean self reported experience of the participants was $2.2\pm0.2$, with range 1 to 5. When reporting experience values the standard error in the mean is presented. A total of 14 (58.4\%) participants were given suggestions from the GA and 10 (41.6\%) were in the control group. The self reported experience of the two groups was comparable, $2.1\pm0.2$ for the GA and $2.2\pm0.4$ for the control. 5 of the 24 participants failed to correctly upload log files following the user study resulting in a total of 19 quantitative data points, of which 11 (58\%) were given level suggestions by the GA and 8 (42\%) were in the control group. When analysing the quantitative data Welch's t-test was used to determine statistically significant deviations between the means of the two groups, p-values of less than 0.05 were considered statistically significant.

\subsection{Results}

\begin{table}[!t]
\renewcommand{\arraystretch}{1.3}
\caption{Q1: Describe the Process you Took to Design a New Level}
\label{fig_design}
\centering
\begin{tabular}{|p{1.6in}||c|c|c|}
\hline
Code & Control & GA & Experience \\
\hline
\multicolumn{4}{|l|}{\emph{Thoughts relating to level design approach}}\\
\hline
Considered player experience/game mechanics & 8 (80\%) & 12 (86\%) & $2.3\pm0.3$ \\
Creating Risk-Reward Trade-off/Balance & 3 (30\%) & 8 (57\%) & $2.4\pm0.3$\\
Encourage/reward exploration & 3 (30\%) & 8 (57\%) & $2.3\pm0.3$\\
Focused on the path from entrance to exit & 4 (40\%) & 6 (43\%) & $1.8\pm0.1$\\
Creating interesting decisions for the player & 4 (40\%) & 6 (43\%) & $2.2\pm0.3$\\
Incremental complexity/difficulty & 2 (20\%) & 4 (29\%) & $2.0\pm0.5$\\
Considered visual aesthetics & 4 (40\%) & 1 (7\%) & $2.6\pm0.5$\\
Aimed to create diversity & 0 (0\%) & 3 (21\%) & $2.7\pm0.5$\\
Used prior experience & 1 (10\%) & 1 (7\%) & $1.5\pm0.4$\\
Unstructured approach & 1 (10\%) & 1 (7\%) & $1.5\pm0.4$\\
\hline
\multicolumn{4}{|l|}{\emph{Thoughts relating to the system/tool}}\\
\hline
Tweaked/edited suggestions from the system & 1 (10\%) & 3 (21\%) & $2.5\pm0.4$\\
Not satisfied by the suggested levels & 0 (0\%) & 1 (7\%) & $2.0\pm0.0$\\
Used suggestions from the system & 1 (10\%) & 0 (0\%)& $2.0\pm0.0$\\
\hline
\end{tabular}
\end{table}

\begin{table}[!t]
\renewcommand{\arraystretch}{1.3}
\caption{Q2: Was Designing 5 Levels Challenging?}
\label{fig_challenge}
\centering
\begin{tabular}{|p{1.6in}||c|c|c|}
\hline
Code & Control & GA & Experience\\
\hline
\multicolumn{4}{|l|}{\emph{Comments related to challenge}}\\
\hline
It was challenging to design multiple levels  & 4 (40\%) & 6 (43\%) & $1.9\pm0.3$\\
It was easy to produce lots of maps  & 3 (30\%) & 7 (50\%) & $1.9\pm0.3$\\
The designs I created ended up similar  & 1 (10\%) & 2 (14\%) & $2.0\pm0.5$\\
\hline
\multicolumn{4}{|l|}{\emph{Comments related to tool/system}}\\
\hline
The tool was useful/helped  & 3 (30\%) & 2 (14\%) & $2.2\pm0.4$\\
The levels generated by the system changed my approach  & 1 (10\%) & 1 (7\%) & $3.0\pm0.7$\\
The tool made it difficult  & 0 (0\%) & 1 (7\%) & $4.0\pm0.0$\\
\hline
\multicolumn{4}{|l|}{\emph{Comments related to the task}}\\
\hline
The limited design space/options made it challenging  & 2 (20\%) & 3 (21\%) & $2.8\pm0.7$\\
It was enjoyable/fun/interesting  & 1 (10\%) & 3 (21\%) & $3.0\pm0.5$\\
The rules of the game were not well defined, so it was difficult  & 2 (20\%) & 1 (7\%) & $2.3\pm0.7$\\
Took longer than expected  & 0 (0\%) & 2 (14\%) & $2.0\pm0.0$\\
\hline
\end{tabular}
\end{table}

\begin{table}[!t]
\renewcommand{\arraystretch}{1.3}
\caption{Q3: Did the Tool Effect the way you Designed your Levels?}
\label{fig_effect}
\centering
\begin{tabular}{|p{1.6in}||c|c|c|}
\hline
Code  & Control & GA & Experience\\
\hline
\multicolumn{4}{|l|}{\emph{Description of the effectiveness}}\\
\hline
It did effect my approach  & 1 (10\%) & 4 (29\%) & $1.8\pm0.2$\\
It moderately effected my approach  & 1 (10\%) & 3 (21\%) & $3.0\pm0.5$\\
It did not effect my approach  & 2 (20\%) & 1 (7\%) & $1.3\pm0.3$\\
\hline
\multicolumn{4}{|l|}{\emph{Discussion of the suggestions presented by tool/system}}\\
\hline
I tweaked suggestions from the system  & 2 (20\%) & 4 (29\%) & $2.0\pm0.2$\\
The suggestions changed my approach  & 2 (20\%) & 4 (29\%) & $3.0\pm0.4$\\
It is good for generating starting points  & 2 (20\%) & 4 (29\%) & $2.0\pm0.2$\\
The suggestions seemed random  & 1 (10\%) & 2 (14\%) & $2.3\pm0.7$\\
I kept generating maps until something good appeared  & 2 (20\%) & 1 (7\%) & $1.7\pm0.3$\\
Suggestions not varied enough  & 0 (0\%) & 2 (14\%) & $2.0\pm0.0$\\
No suggestions were useful/helpful  & 1 (10\%) & 1 (7\%) & $1.5\pm0.4$\\
I had to significantly modify the suggestions  & 0 (0\%) & 2 (14\%) & $3.0\pm0.7$\\
Suggestions rarely got the treasure/enemy layout right  & 0 (0\%) & 2 (14\%) & $2.5\pm0.4$\\
I tried to influence the suggestions  & 0 (0\%) & 1 (7\%) & $2.0\pm0.0$\\
Some of the generated maps were unsuitable  & 0 (0\%) & 1 (7\%) & $2.0\pm0.0$\\
\hline
\end{tabular}
\end{table}

\begin{table}[!t]
\renewcommand{\arraystretch}{1.3}
\caption{Q4: How Would you Describe the Tool to Someone Else?}
\label{fig_description}
\centering
\begin{tabular}{|p{1.6in}||c|c|c|}
\hline
Code & Control & GA & Experience\\
\hline
\multicolumn{4}{|l|}{\emph{It is a tool which works with the designer}}\\
\hline
It learns from your seed designs & 2 (20\%) & 5 (36\%) & $1.9\pm0.2$\\
It generates starting points - you'll need to edit them  & 1 (10\%) & 4 (29\%) & $1.6\pm0.2$\\
It suggests different levels to you  & 1 (10\%) & 4 (29\%) & $2.4\pm0.6$\\
It helps inspire new ideas  & 1 (10\%) & 1 (7\%) & $1.5\pm0.4$\\
It is a rapid prototyping tool  & 1 (10\%) & 1 (7\%) & $3.0\pm0.7$\\
It is an interactive tool for PCG  & 0 (0\%) & 1 (7\%) & $2.0\pm0.0$\\
\hline
\multicolumn{4}{|l|}{\emph{It is a tool which works independently from the designer}}\\
\hline
It randomly generates levels  & 2 (20\%) & 2 (14\%) & $1.7\pm0.3$\\
No inclusion of human approach to games  & 0 (0\%) & 1 (7\%) & $3.0\pm0.0$\\
\hline
\multicolumn{4}{|l|}{\emph{Description of UI/UX}}\\
\hline
Functional description of UI  & 3 (30\%) & 2 (14\%) & $1.4\pm0.2$\\
It is fun/enjoyable  & 1 (10\%) & 1 (7\%) & $1.5\pm0.4$\\
The tool can be tedious  & 0 (0\%) & 1 (7\%) & $2.0\pm0.0$\\
\hline
\end{tabular}
\end{table}
Tables~\ref{fig_design} to~\ref{fig_description} show the codes and frequencies for all the qualitative questions, along with the mean self reported experience (as described in \ref{mat}) of participants who responded with each code. When answering Q1 participants predominately (N=20, 83.3\%) reported considering the player experience when designing levels. For example, \emph{``The first level ensured an easy layout where everything is encountered and choice is allowed..."}. There were some notable differences between the ways the two groups answered this question. More participants in the GA group were interested in creating levels which rewarded and encouraged exploration (57\% compared to 30\%), and creating a diverse set of levels (21\% compared to 0\%). The participants focusing on diversity tended to be those with more design experience. Furthermore, more participants in the GA group reported tweaking suggestions from the system (21\% compared to 10\%).

When answering Q2 50\% of the GA group described the task as easy, compared to 30\% of the control group. 14\% of the GA group reported that the tool was helpful, compared to 30\% of the control group. However, when answering Q3, 50\% of the GA group reported that the tool had an effect on their design approach compared to 20\% of the control group. 

\begin{figure}[!t]
\centering
\includegraphics[width=2in]{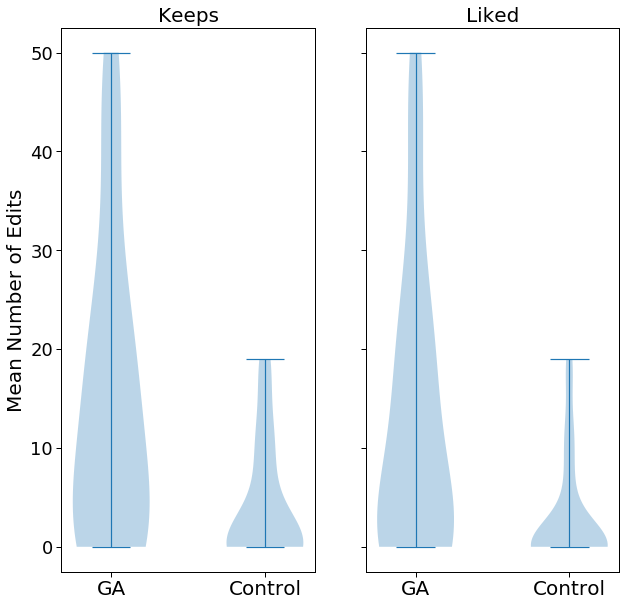}
\caption{Comparing the number of edits of liked and kept levels between the GA and control group. The differences in the means of these distributions is statistically significant (p-value $<0.01$).}
\label{fig_edits}
\end{figure}
The only statistically significant difference between the two groups in the quantitative data was the number of edits of suggestions from the system. The mean number of edits of liked maps was 13.00 (SD = 14.59) and 2.82 (SD = 5.34) for the GA and control groups respectively. For kept maps the mean number of edits was 14.81 (SD = 14.89) for GA, and 4.10 (SD = 6.02) for the control group. In both cases the p-value was less than 0.01, the full distributions are shown in Figure~\ref{fig_edits}. When put in the context of our qualitative findings, which suggest that the participants in the GA group were more engaged with the task, we can interpret this increased number of edits as increased engagement. When answering Q3, participants in the GA group were more likely to discuss tweaking suggestions, how the suggestions changed their approach and how the suggestions were good starting points (29\% compared to 20\% for all three responses). Participants who reported that the suggestions changed their approach had more design experience. When discussing the level of challenge of the task, 21\% of the GA group described the task as enjoyable compared to 10\% of the control group, these participants tended to have more design experience. When answering Q2, 14\% of the GA group reported that the task took longer than expected, compared to 0\% of the control group. This could be evidence of increased engagement, but it was unclear if the participants saw this positively or negatively. There was a trend that participants in the GA group took more iterations to design 5 levels, but there is not enough data to show statistical significance. 
\begin{figure}[!t]
\centering
\includegraphics[width=2in]{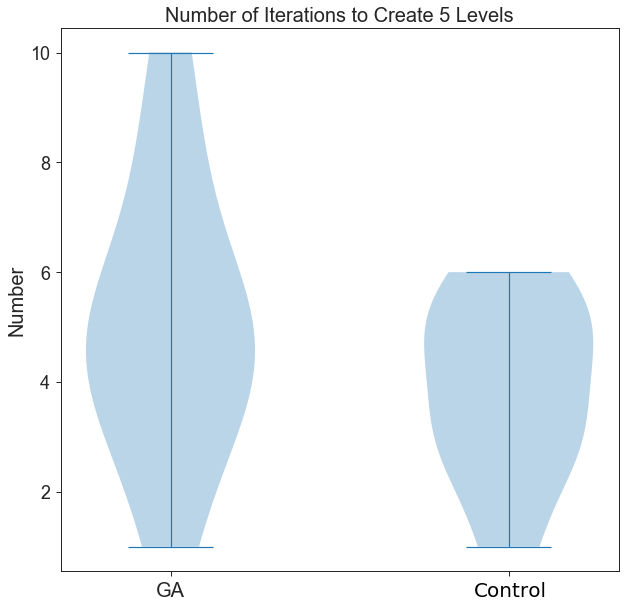}
\caption{Comparing the number of iterations taken to design 5 levels, when users were given level suggestions by the GA to random suggestions (p-value = 0.22)}
\label{fig_iterations}
\end{figure}
The mean number of iterations taken to design 5 levels by the GA group was 5.08 (SD = 2.39) and the control group 3.88 (SD = 1.53), the distribution is plotted in Figure~\ref{fig_iterations}.

The mean number of likes per iteration for GA group was 1.31 (SD = 1.24) and control group 1.84 (SD = 1.15). The p-value for this comparison was 0.71 meaning that there was no statistical significance. In the group of participants who were presented suggestions by the GA there were 2 cases where a user used the blank canvas to create a new design from scratch, and 1 case in the control group. We can not conclude a statistically significant difference from this data. This data is plotted in Figure~\ref{fig_time_history}.
\begin{figure*}[!t]
\centering
\subfloat[Number of Likes]{\includegraphics[width=3in]{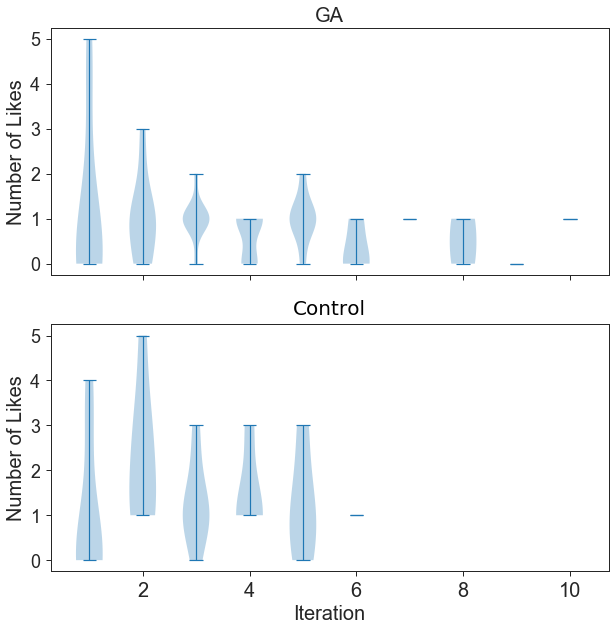}}
\hfil
\subfloat[Number of Keeps]{\includegraphics[width=3in]{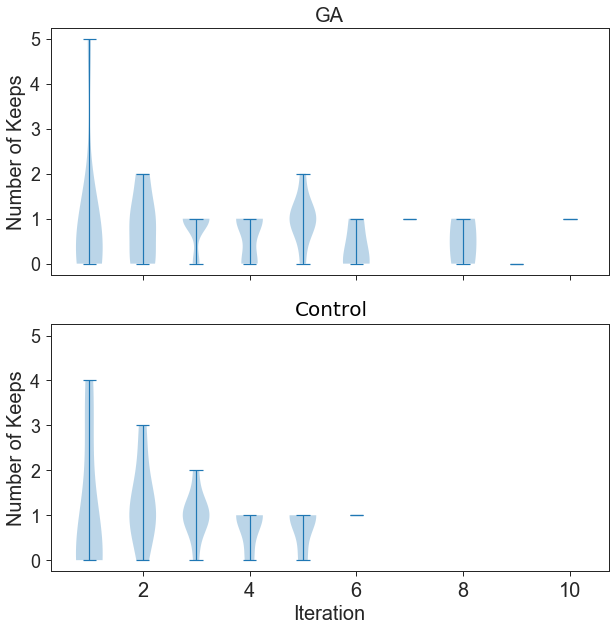}}
\hfil
\caption{Comparing the distribution of the number of Likes and Keeps at each iteration between the GA and Control groups}
\label{fig_time_history}
\end{figure*}

In the final question participants were asked to describe the tool, Table~\ref{fig_description} shows the results from coding the answers to this question. There were two identifiable groups of description based on how a level designer interacts with the tool. Participants either described it as a tool which works with or independently to the designer. More participants from the GA group described the tool as something which works with the designer. 36\% of the GA group described the tool as learning from your designs compared to 20\% from the control group. Interestingly, 29\% of the GA group stated that the tool generates starting points which you are required to edit, compared to 10\% of the control group. This puts the responses for Q3 into context, which suggest that the GA group were more engaged with modifying the suggestions from the system. 14\% of the GA group described the tool as something which randomly generates levels, compared to 20\% from the control group. Participants from the GA group were less likely to simply describe the UI features of the tool than those from the control group (14\% compared to 30\%).



\section{Discussion}

A common thread throughout the qualitative data was that those participants who were given suggestions by the GA talked a lot more about the suggestions the system gave them. They described the tool as learning from the designs they created and as a tool which works with the designer to support prototyping. The participants from the control group focused more on the functional description of the UI and generally provided less detailed responses to questions. This general lack of engagement from participants in the control group is further supported by the quantitative data which showed that the GA group edited suggestions by the system more than the control group. Initially we thought this was an indication that the GA was doing a bad job, but when taken in context of the qualitative data we found that these participants were considering their designs much more---as one participant stated, the suggestions sparked new ideas. This highlights that a mixed-methods approach is essential when evaluating mixed-initiative systems, quantitative data only tells half the story. The engagement narrative is further supported by the answers to Q1 where participants from the GA group were more likely to consider higher level design concepts, such as rewarding exploration, in their design process. There was also a general trend that the GA group took more iterations to complete the task. We were surprised to find that as many participants in the control group described the suggestions as being useful as in the GA group, suggesting that any suggestions are helpful to the creative process. When comparing the self reported experience levels of participants we found that more experienced participants (a) were more concerned about diversity, (b) reported that the tool changed their approach and (c) enjoyed the task. Overall our data shows that the system we designed does support designers through the design process and is more effective than random suggestions.

\subsection{Evaluation of Our Scientific Approach}
In hindsight it would have been appropriate to have included some Likert scale questions as part of our user survey. In particular, with Q3 we found that not all participants clearly stated if the tool affected their approach, which would have been captured by a scale response. Performing the study on-line introduces problems such as not all participants correctly submitting log files and possible minor differences in experience based on the hardware they are running. A larger group of participants would have resulted in the generation of more or different codes during the thematic analysis, but that does not mean those codes would have been better. The objective of thematic analysis is not to determine all possible themes, but to generate themes based on the data collected, and to use those to identify patterns of meaning to answer research questions~\cite{Braun2019-kt, Braun2019-iv}. The patterns we identified align well with other research in this field and add further evidence that mixed-initiative tools can support the creative process. 

\subsection{Future Work}
One limitation of the tool was that a number of participants in the GA group noted that the suggestions which were given were all too similar to each other. In the future it would be interesting to add more sophisticated mechanisms~\cite{Preuss2014-yh, Sampaio2017-mu, Melotti2019-rt, Alvarez2019-oc} to ensure diversity in the suggestions. A further line of enquiry would be to take a model based approach as explored by Alvarez et al.~\cite{Alvarez2020-xv} and build a model to predict which maps the designer has a preference for. We suggest that Bayesian Optimisation would be a good avenue to explore in addition to machine learning, this is a model based approach that builds a surrogate model of the function it is optimising. In this application that surrogate model could be of the designers design preference.

\section*{Acknowledgment}

The authors would like to thank Stephen Lindsay for his excellent advice for designing the user study, the participants for donating their time to our work, and the reviewers for offering constructive valuable feedback.

\ifCLASSOPTIONcaptionsoff
  \newpage
\fi



\bibliographystyle{IEEEtran}
\bibliography{IEEEabrv,papers}
%



%

\begin{IEEEbiography}[{\includegraphics[width=1in,height=1.25in,clip,keepaspectratio]{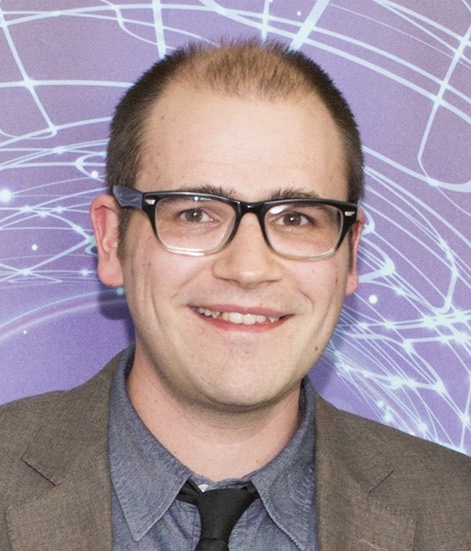}}]{Sean P. Walton} received his MPhys degree in Physics from Aberystwyth in 2002. He worked as a school Physics teacher for several years before completing his PhD in numerical methods at Swansea Universities' Zienkiewicz Centre for Computational Engineering in 2013. Currently he works as a senior lecturer in Computer Science at Swansea University. His academic research focus is on using evolutionary optimisation algorithms to support the design process in a number of fields, and investigating game design approaches that are effective for educational games. Outside of academia he is a BAFTA Cymru nominated game designer and founding director of Pill Bug Interactive.
\end{IEEEbiography}

\begin{IEEEbiography}[{\includegraphics[width=1in,height=1.25in,keepaspectratio, trim={55mm 80mm 50mm 0}, clip=true]{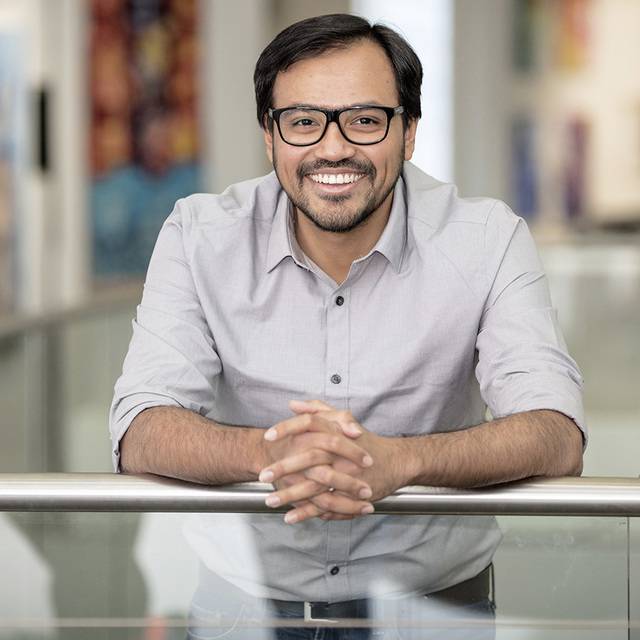}}]
{Alma A. M. Rahat} is a Lecturer in Big Data/Data Science at Swansea University, UK. He has a BEng (Hons) in Electronic Engineering from the University of Southampton, UK, and a PhD in Computer Science from the University of Exeter, UK. He worked as a product development engineer after his bachelor's degree, and held post-doctoral research positions at the University of Exeter. Before moving to Swansea, he was a Lecturer in Computer Science at the University of Plymouth. His current research focus is in the broad areas of fast hybrid optimisation methods, real-world problems and machine learning. In particular, he is developing efficient methods inspired from surrogate-assisted (Bayesian) optimisation for optimising computationally or financially expensive problems (for example, computational fluid dynamics aided design problems).
\end{IEEEbiography}

\begin{IEEEbiography}[{\includegraphics[width=1in,height=1.25in,clip,keepaspectratio]{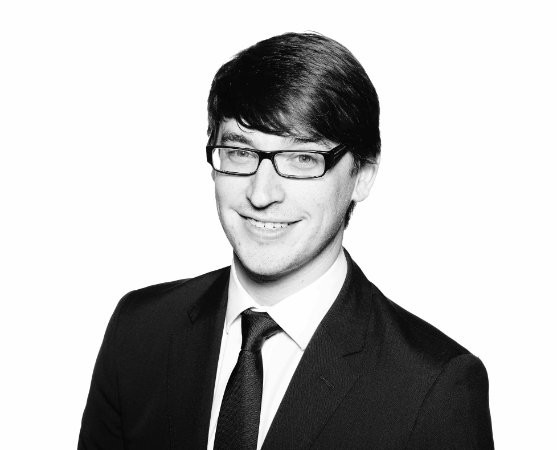}}]{James Stovold} received his MEng degree in Computer Science from York in 2012, and completed his PhD at the York Cross-Disciplinary Centre for Systems Analysis (YCCSA) in 2016. After a brief stint in industry, he returned to teaching in 2018. Currently a lecturer at the British University Vietnam, his interests are in distributed cognition, bio-inspired algorithms, swarm intelligence, and robotics.
\end{IEEEbiography}






\fi

\iftrue

\newpage
\onecolumn
\setcounter{page}{1}
\section*{Supplementary Materials}

\subsection{Algorithm-Driven Benchmarks} 

\subsubsection{Methodology} \label{sec:tuning}
To test the effectiveness of the GA itself a series of studies were performed using an entirely algorithm driven-approach. For benchmark tests we use six maps presented by Baldwin et al.~\cite{Baldwin2017-me} to show the different styles of maps which could be created by varying the configurations of their approach. Conveniently they represent a range of styles, from maps with no corridors to maps with no chambers. We use these maps as benchmarks to avoid unconscious bias which could result from us designing our own. For each test the target map was entered as the initial user designed level. The GA was then run and the highest ranked map in the population at the end of the optimisation is presented. 

\subsubsection{Results}
\begin{figure}[!t]
\centering
\subfloat[Initial Design]{\includegraphics[width=1.5in]{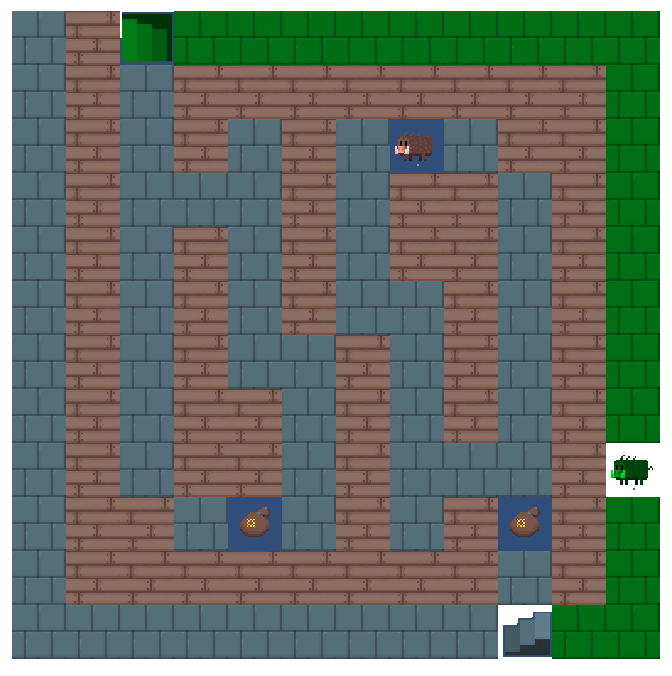}\label{fig_algo_A_T}} 
\hfil
\subfloat[Output]{\includegraphics[width=1.5in]{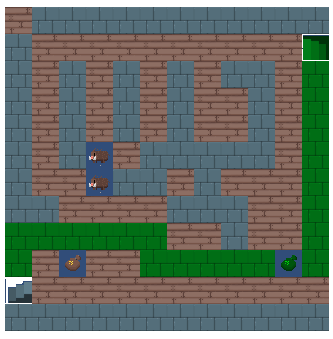}\label{fig_algo_A_O}} 
\hfil
\subfloat[Sum of fitness functions for the highest ranked individual each generation, and mean of population]{\includegraphics[width=3in]{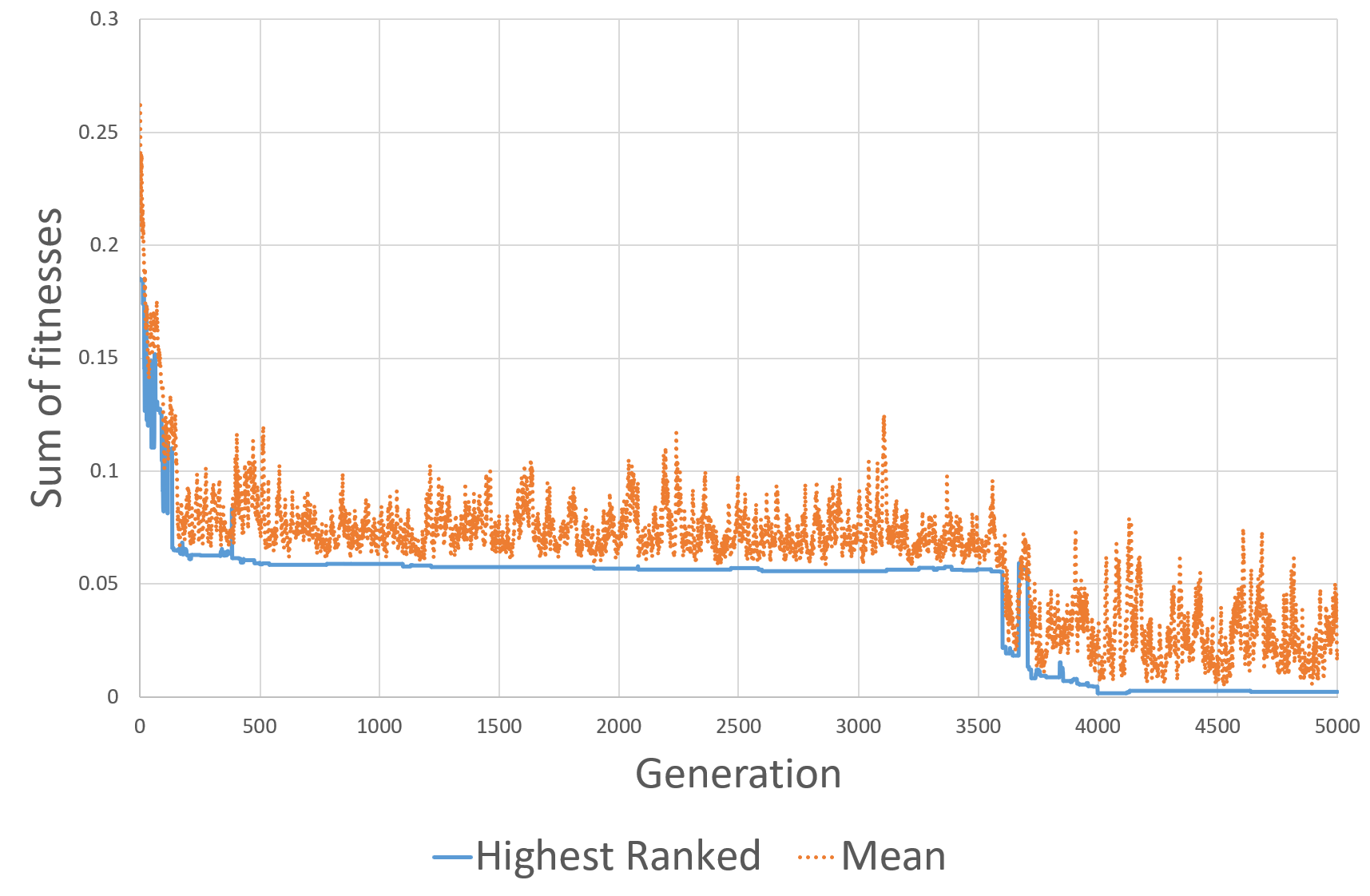}\label{fig_algo_A_G}} 
\hfil
\caption{Algorithm-Driven Test A.}
\label{fig_algo_A}
\end{figure}
The first-algorithm driven test is one which is made up of corridors with zero chambers. The target map is shown in Figure~\ref{fig_algo_A_T}, and the map created by the GA is shown in Figure~\ref{fig_algo_A_O}. The created map contains only one chamber, is predominately made up of corridors and has the same number of treasure and enemy tiles as the targets. For this first test we have included an optimisation history graph, Figure~\ref{fig_algo_A_G}, constructed by taking the sum of fitnesses for the best individual each generation. It is typical of the behaviour observed in all tests.
\begin{figure}[!t]
\centering
\subfloat[Initial Design]{\includegraphics[width=1.5in]{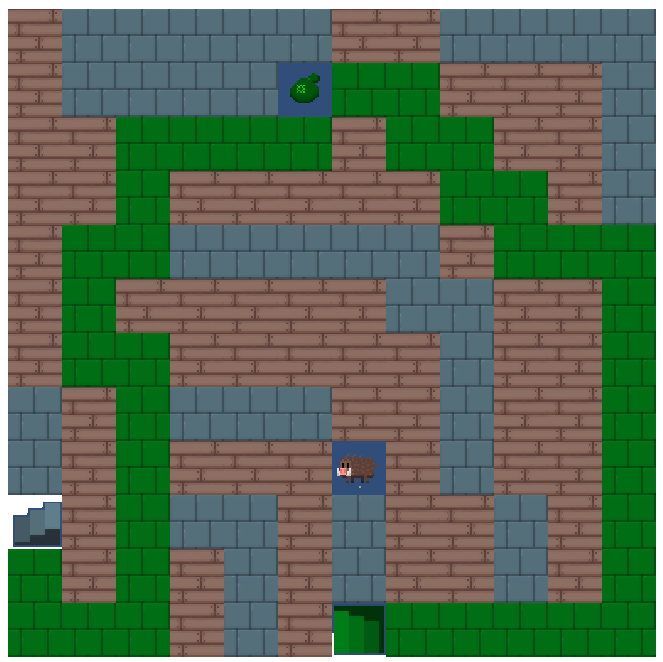}}
\hfil
\subfloat[Output]{\includegraphics[width=1.5in]{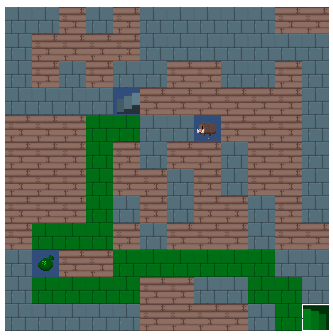}}
\hfil
\caption{Algorithm-Driven Test B}
\label{fig_algo_B}
\end{figure}
The results from test B are shown in Figure~\ref{fig_algo_B}. In this test the target map has a single chamber and many corridors. The resulting output is dominated by corridors, although some of them are unreachable by the player. The output design has a similar ratio of passable to impassible tiles. Both maps have a single treasure and enemy tile.
\begin{figure}[!t]
\centering
\subfloat[Initial Design]{\includegraphics[width=1.5in]{5c_in.png}}
\hfil
\subfloat[Output]{\includegraphics[width=1.5in]{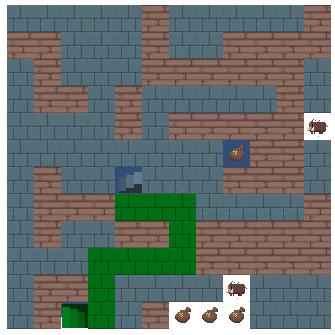}}
\hfil
\caption{Algorithm-Driven Test C}
\label{fig_algo_C}
\end{figure}
Test C is a map with a comparable number of corridors to chambers. The results of this study are shown in Figure~\ref{fig_algo_C}. The output has a similar balance of corridors and chambers, and a similar distribution of treasures and enemies.
\begin{figure}[!t]
\centering
\subfloat[Initial Design]{\includegraphics[width=1.5in]{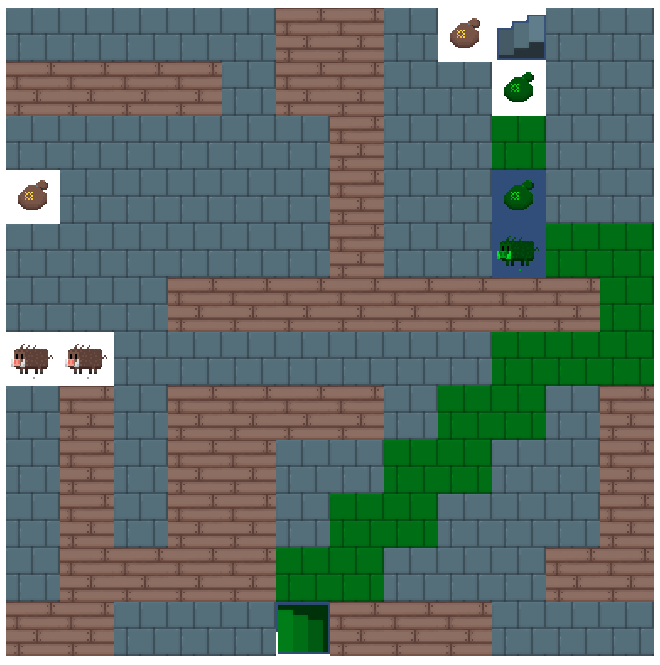}}
\hfil
\subfloat[Output]{\includegraphics[width=1.5in]{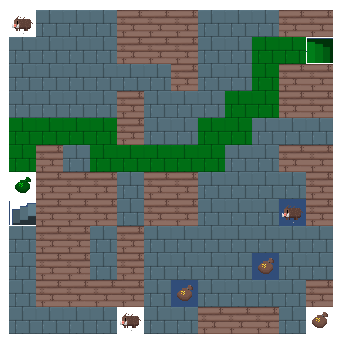}}
\hfil
\caption{Algorithm-Driven Test D}
\label{fig_algo_D}
\end{figure}
The results for test D are shown in Figure~\ref{fig_algo_D}. This target design is largely made up of chambers with a few corridors. The GA is capable of matching this distribution. 
\begin{figure}[!t]
\centering
\subfloat[Initial Design]{\includegraphics[width=1.5in]{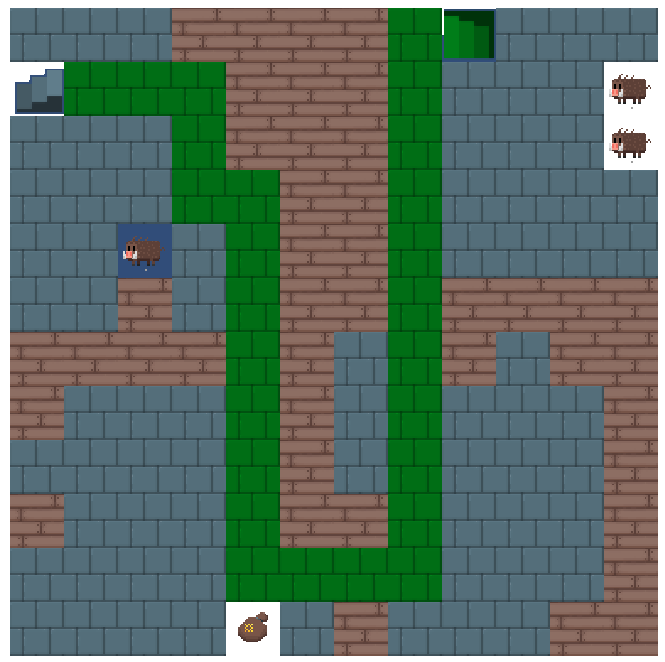}}
\hfil
\subfloat[Output]{\includegraphics[width=1.5in]{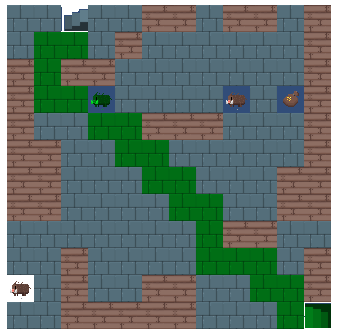}}
\hfil
\caption{Algorithm-Driven Test E}
\label{fig_algo_E}
\end{figure}
In test E the target map is made up of chambers connected by single tile corridors. The results of this test are shown in Figure~\ref{fig_algo_E}. Much like the target the output is made up of chambers and single tile corridors with the same number of treasure and enemy tiles.
\begin{figure}[!t]
\centering
\subfloat[Initial Design]{\includegraphics[width=1.5in]{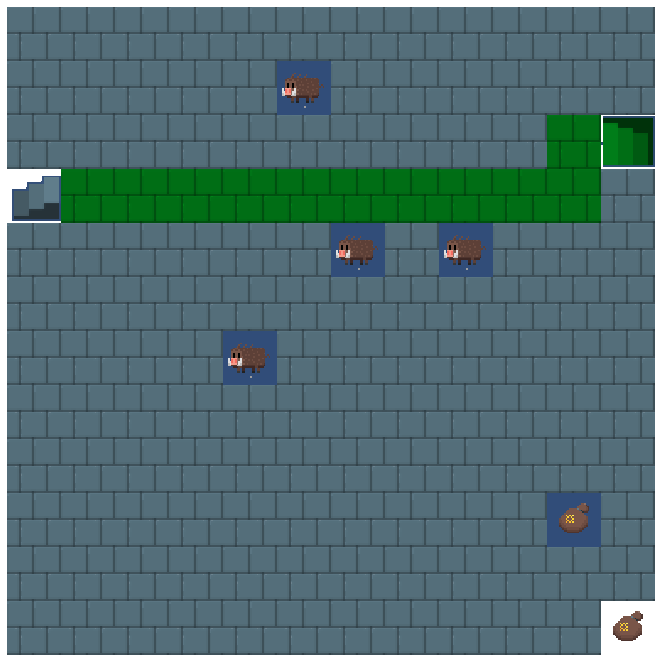}}
\hfil
\subfloat[Output]{\includegraphics[width=1.5in]{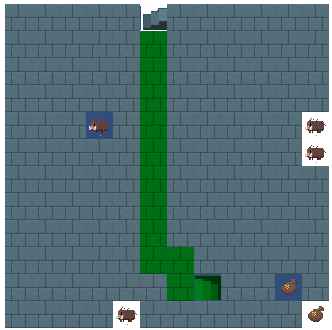}}
\hfil
\caption{Algorithm-Driven Test F}
\label{fig_algo_F}
\end{figure}
The final test, F, is simply a map with zero wall tiles. Figure~\ref{fig_algo_F} shows that the GA handles this edge case. Also notice that the path length is almost the same in both.

\subsection{Final Designs of Participants}
The following images show the final screen after participants had completed their task. As a reminder, the top 5 maps are the participants final 5 levels. The larger maps are the suggestions from the system which may have been modified by the user before this image was taken.

\subsubsection{Control Group}
Figures~\ref{ConA} to~\ref{ConG} show the final screens for participants in the control group.
\begin{figure}[!t]
    \centering
    \includegraphics[width=3.5in]{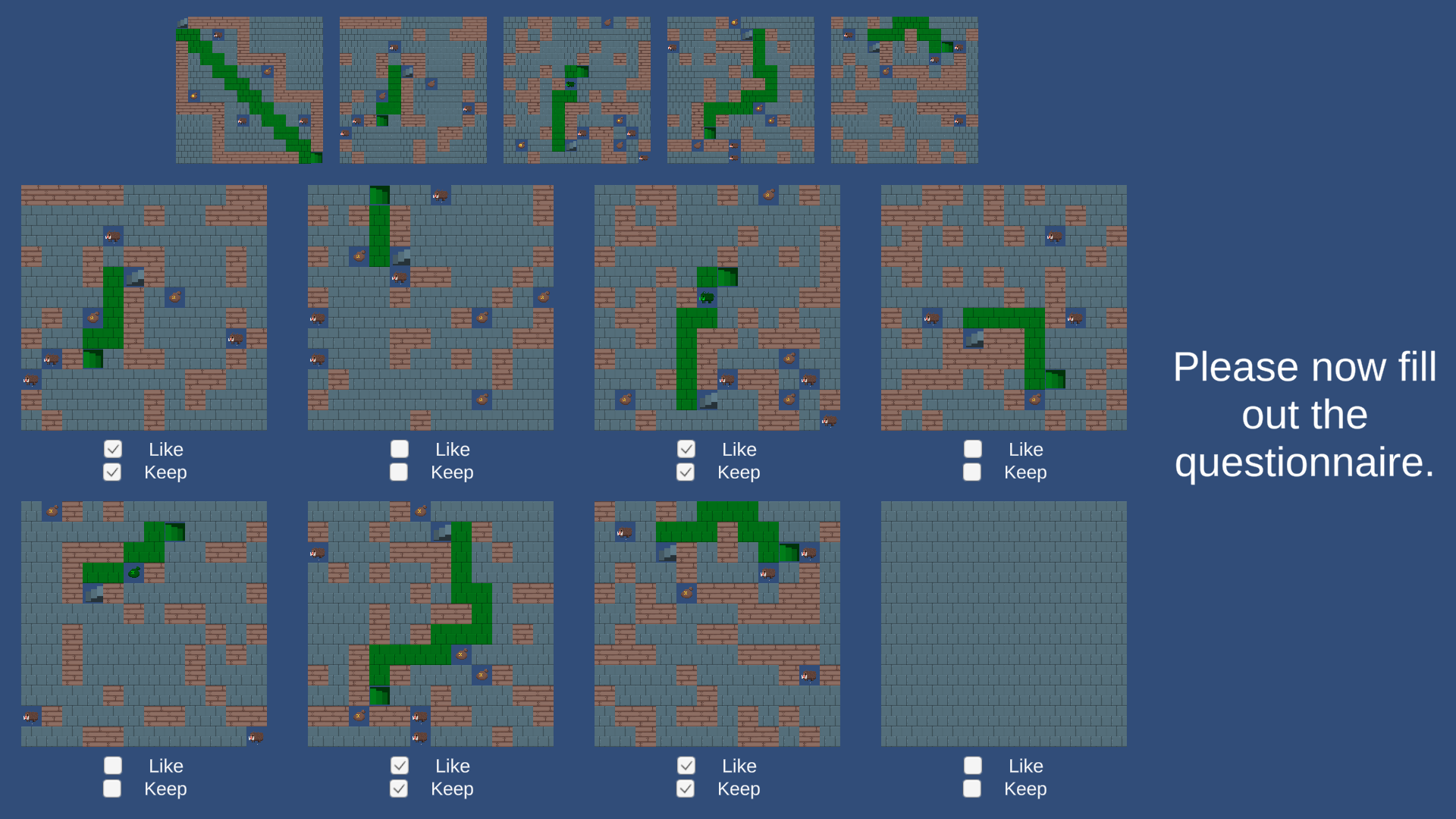}
    \caption{Control Participant A}
    \label{ConA}
\end{figure}
\begin{figure}[!t]
    \centering
    \includegraphics[width=3.5in]{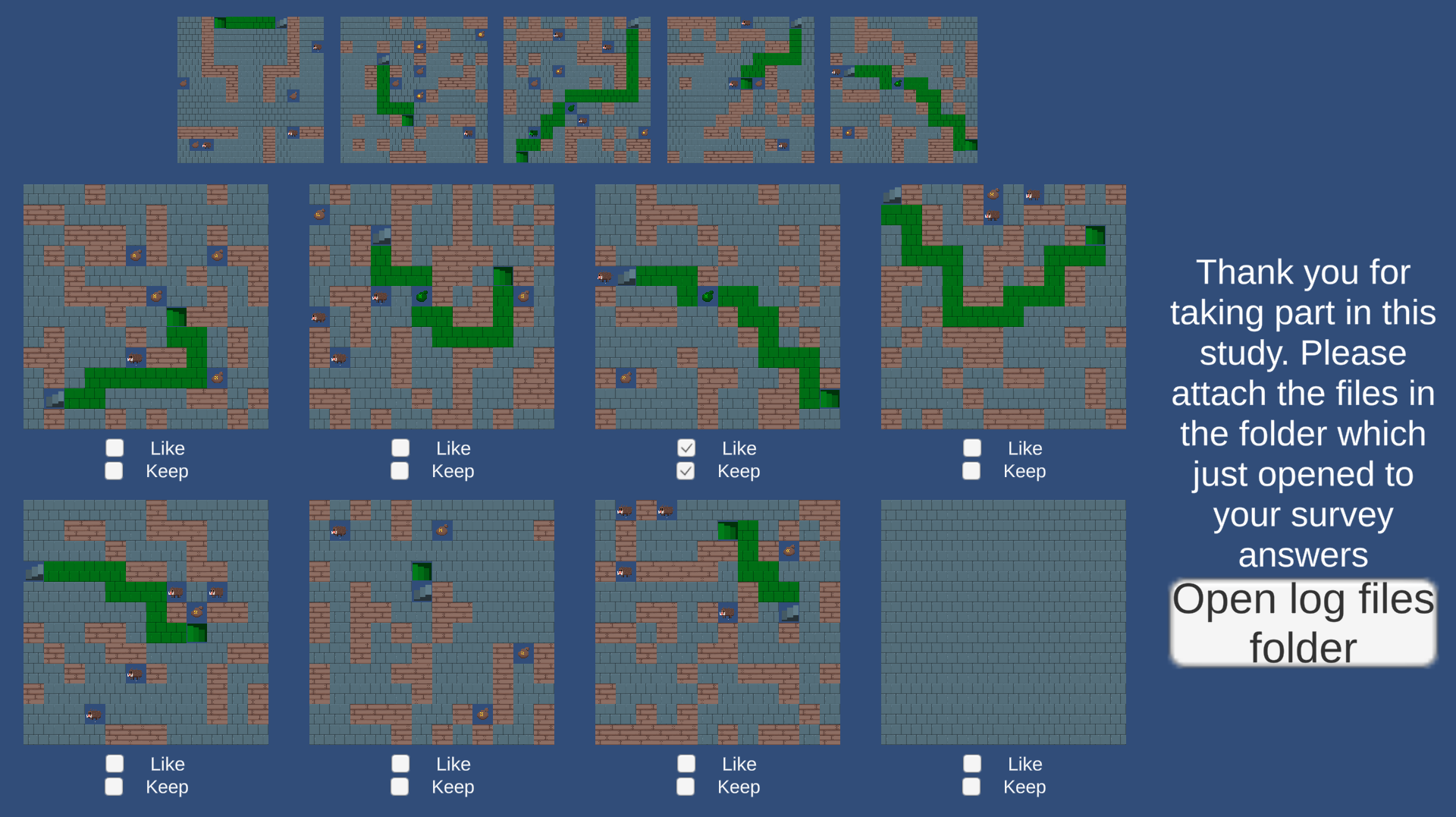}
    \caption{Control Participant B}
\end{figure}
\begin{figure}[!t]
    \centering
    \includegraphics[width=3.5in]{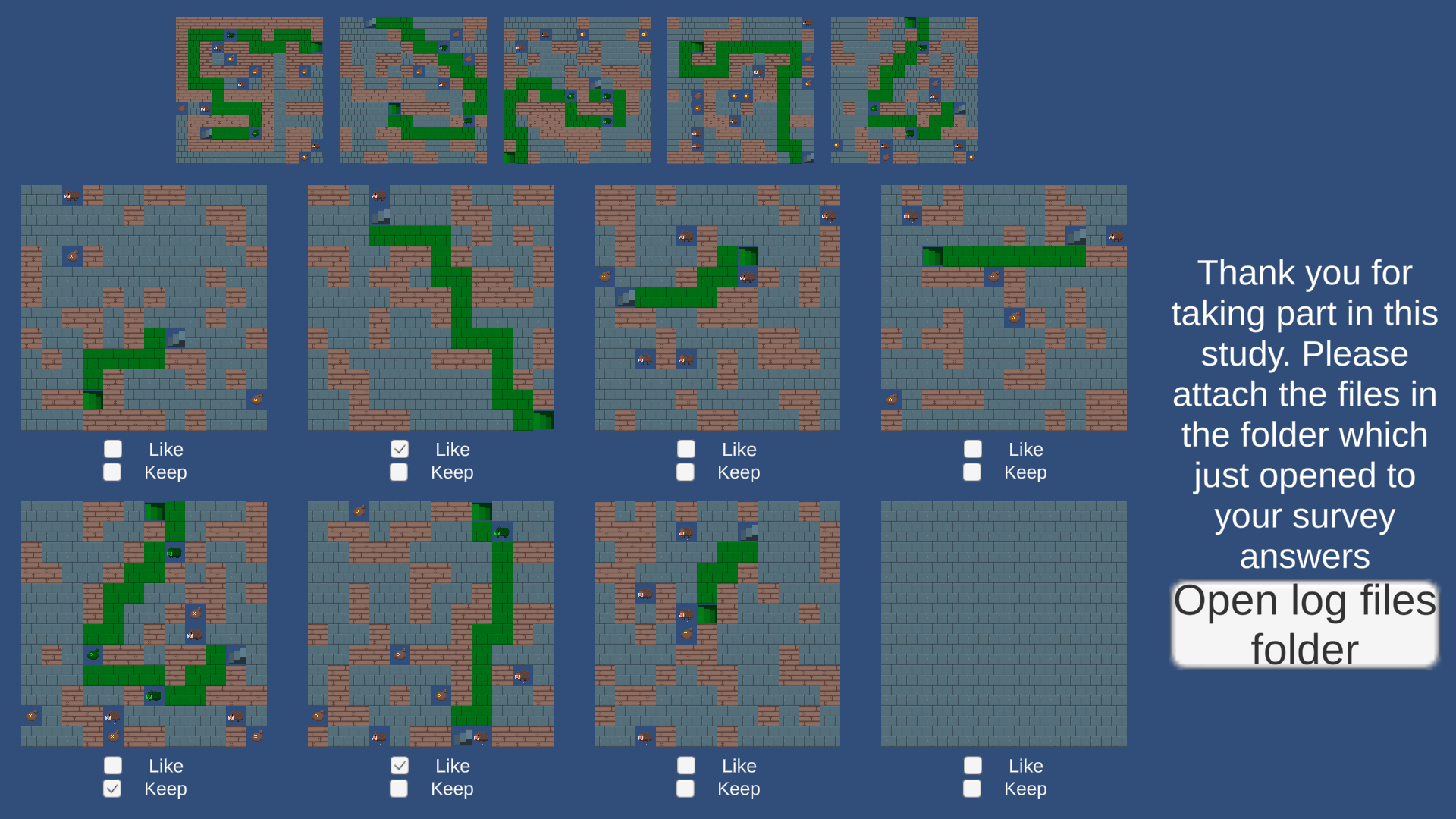}
    \caption{Control Participant C}
\end{figure}
\begin{figure}[!t]
    \centering
    \includegraphics[width=3.5in]{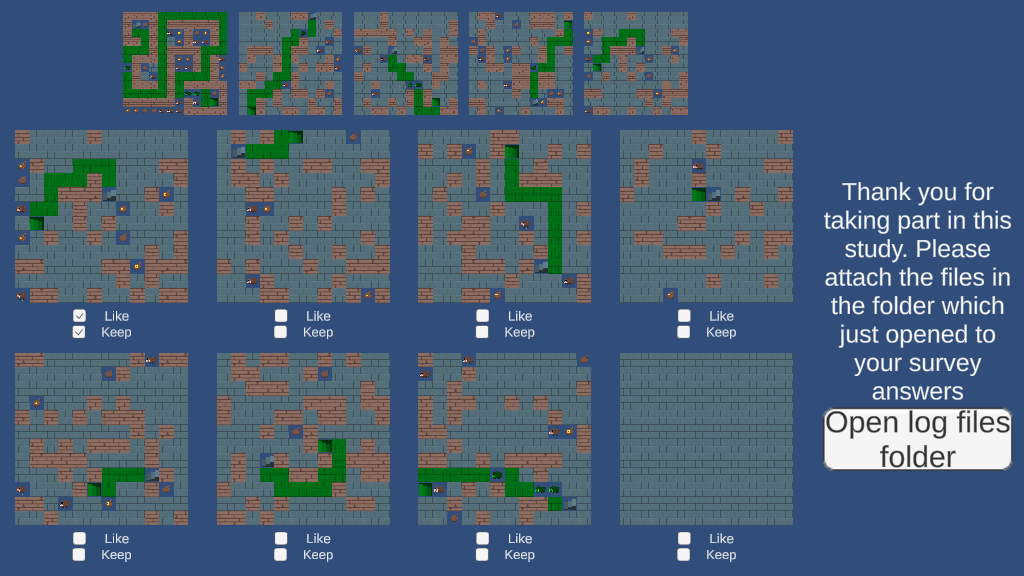}
    \caption{Control Participant D}
\end{figure}
\begin{figure}[!t]
    \centering
    \includegraphics[width=3.5in]{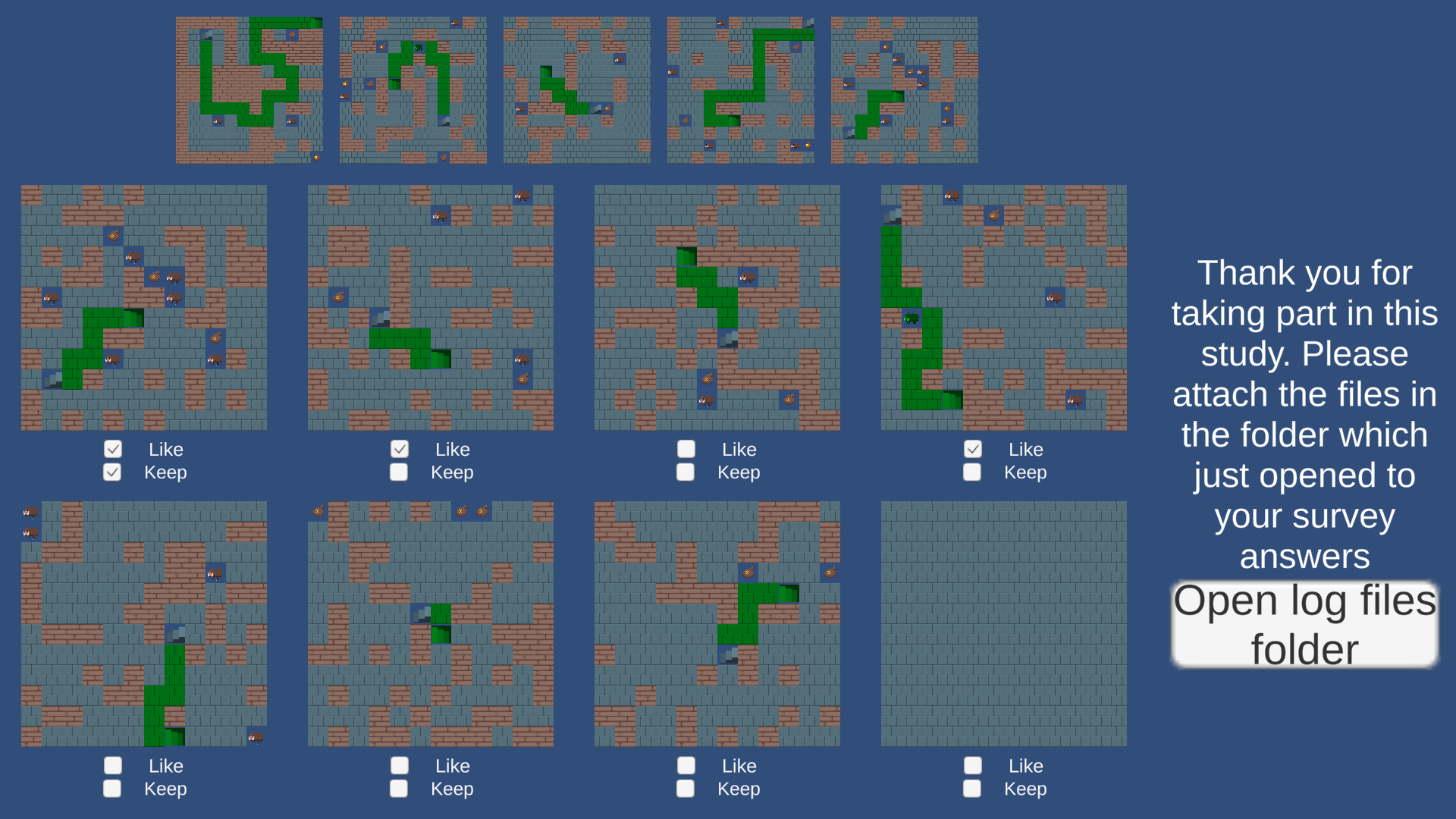}
    \caption{Control Participant E}
\end{figure}
\begin{figure}[!t]
    \centering
    \includegraphics[width=3.5in]{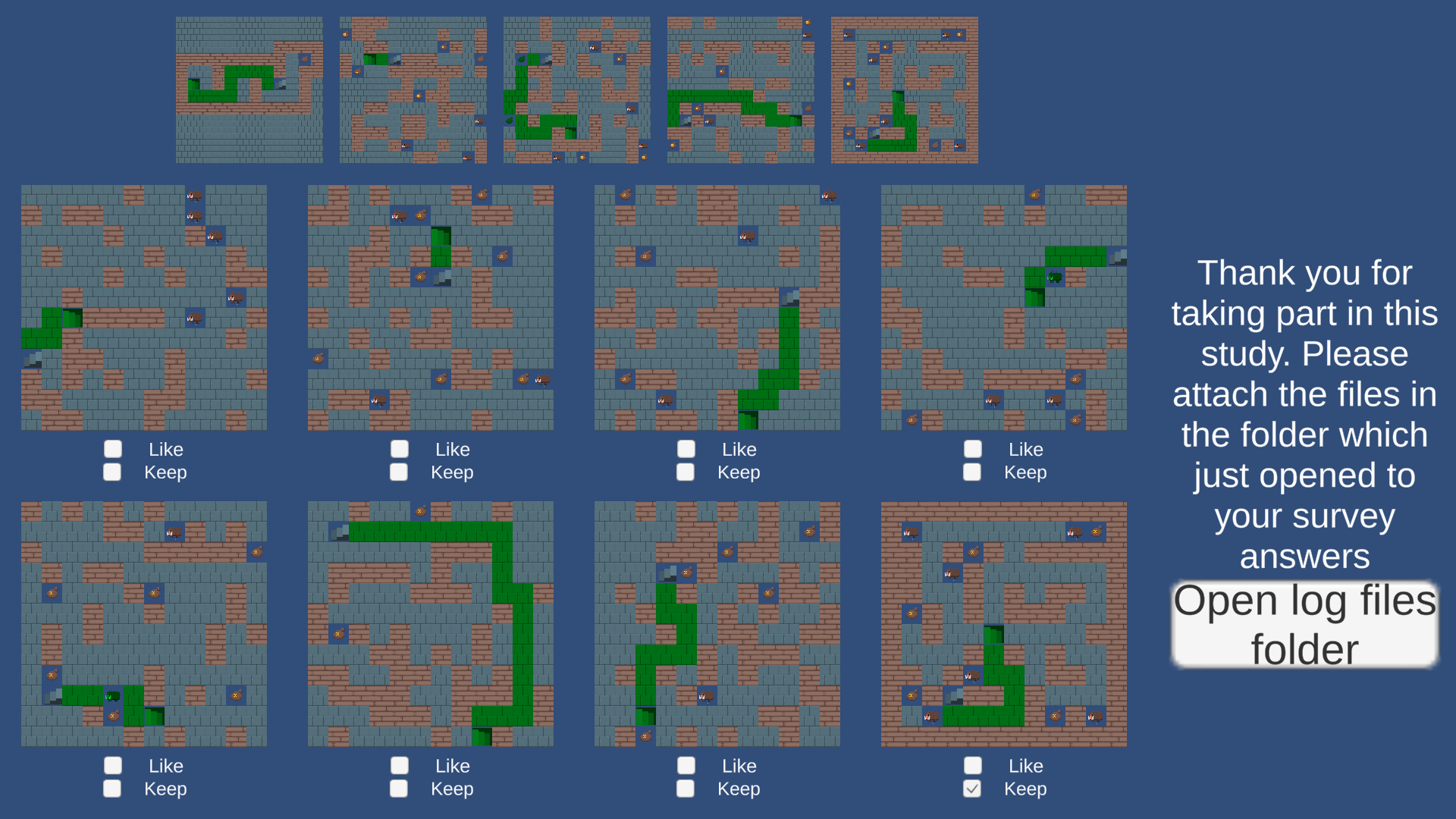}
    \caption{Control Participant F}
\end{figure}
\begin{figure}[!t]
    \centering
    \includegraphics[width=3.5in]{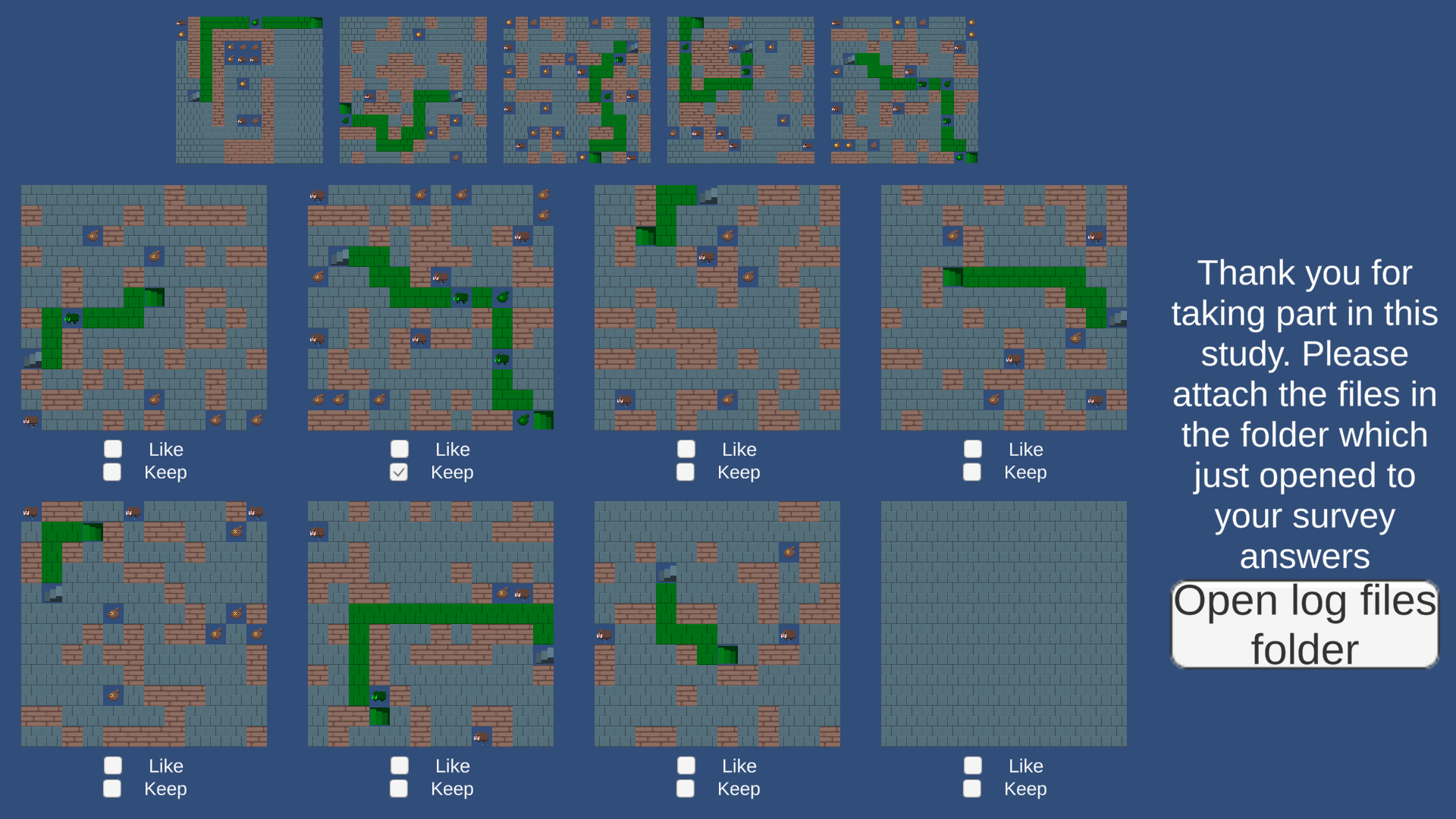}
    \caption{Control Participant G}
    \label{ConG}
\end{figure}

\subsubsection{GA Group}
Figures~\ref{GAA} to~\ref{GAK} show the final screens for participants in the genetic algorithm group.
\begin{figure}[!t]
    \centering
    \includegraphics[width=3.5in]{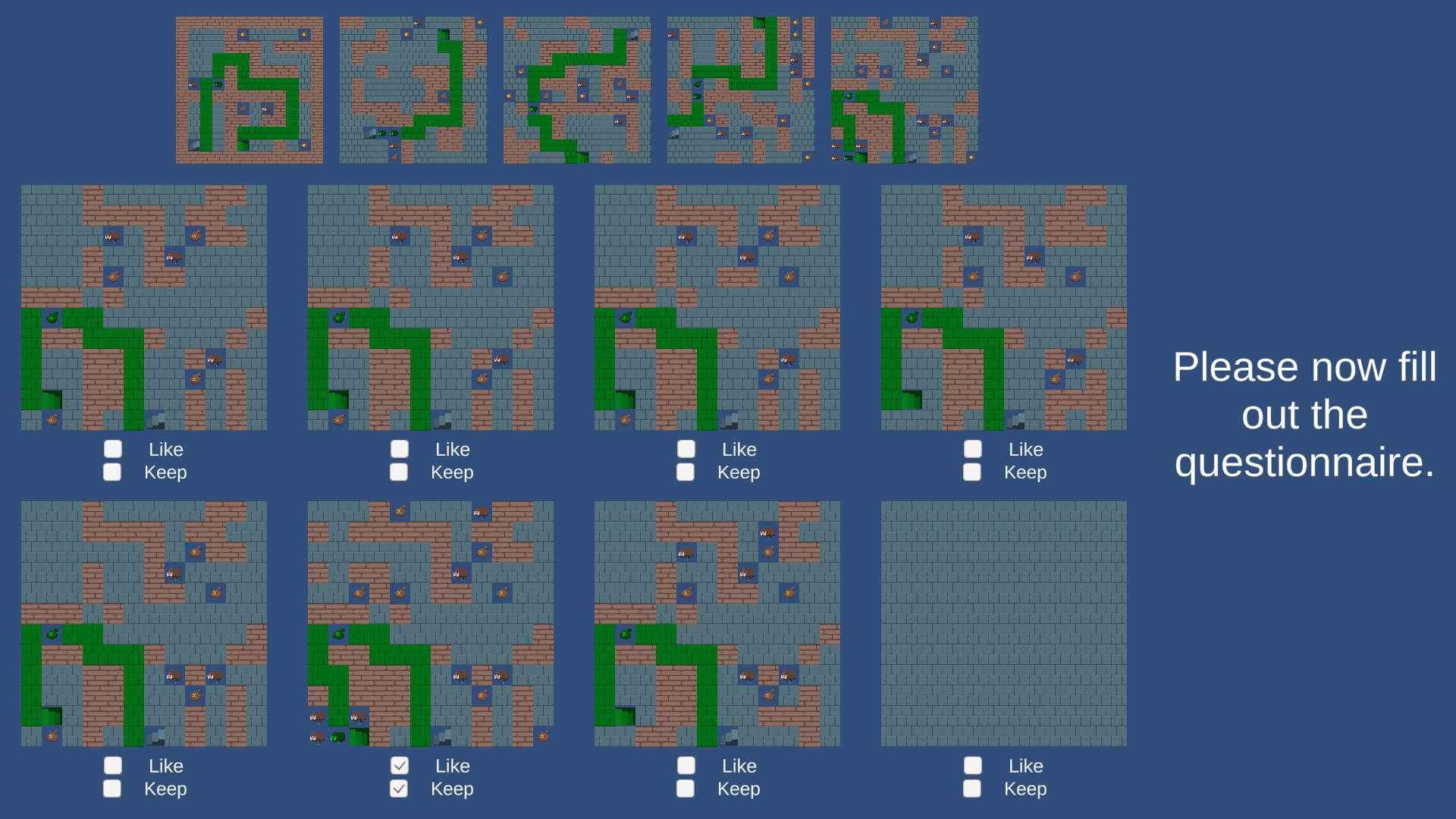}
    \caption{Genetic Algorithm Participant A}
    \label{GAA}
\end{figure}
\begin{figure}[!t]
    \centering
    \includegraphics[width=3.5in]{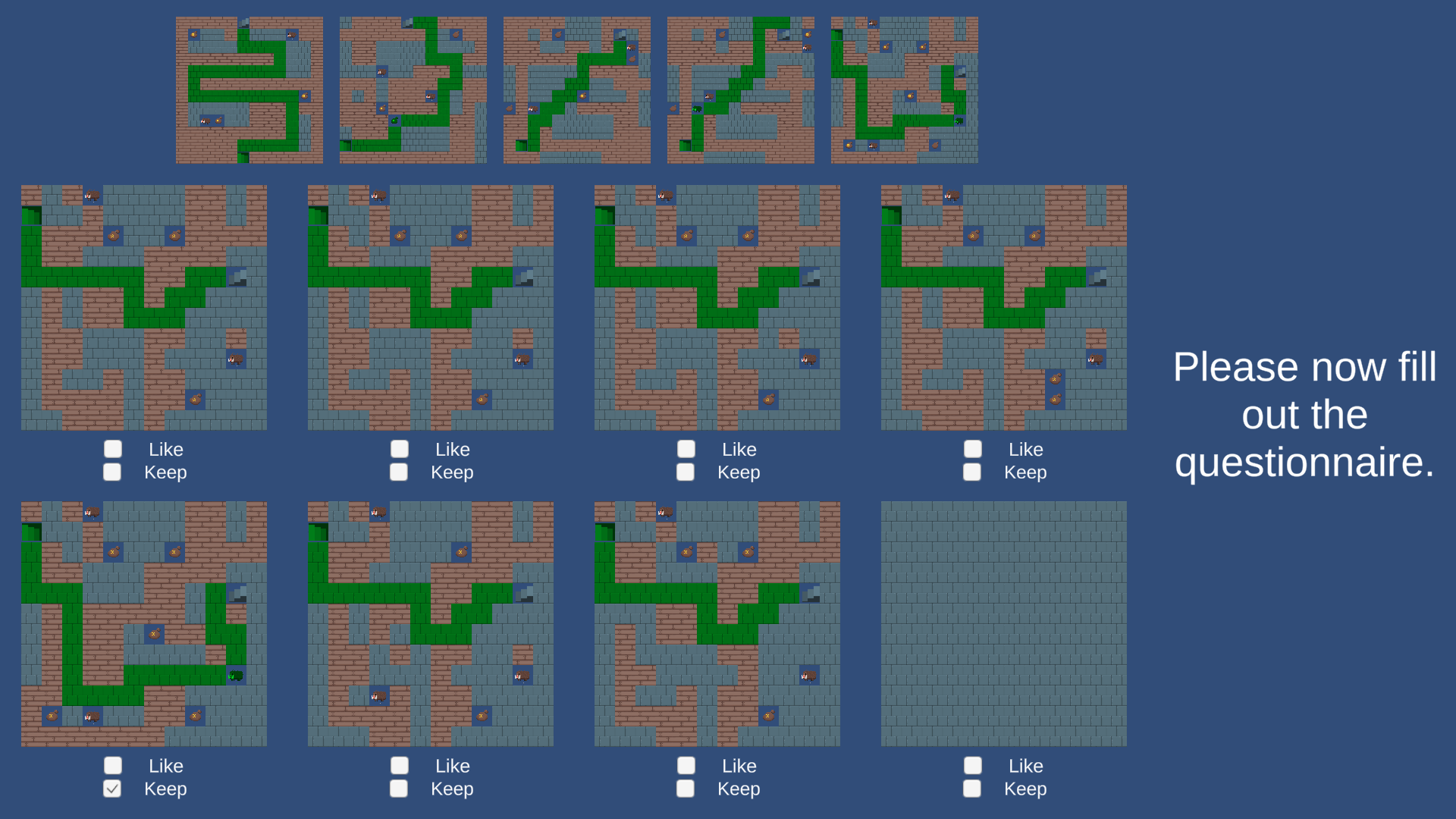}
    \caption{Genetic Algorithm Participant B}
    
\end{figure}
\begin{figure}[!t]
    \centering
    \includegraphics[width=3.5in]{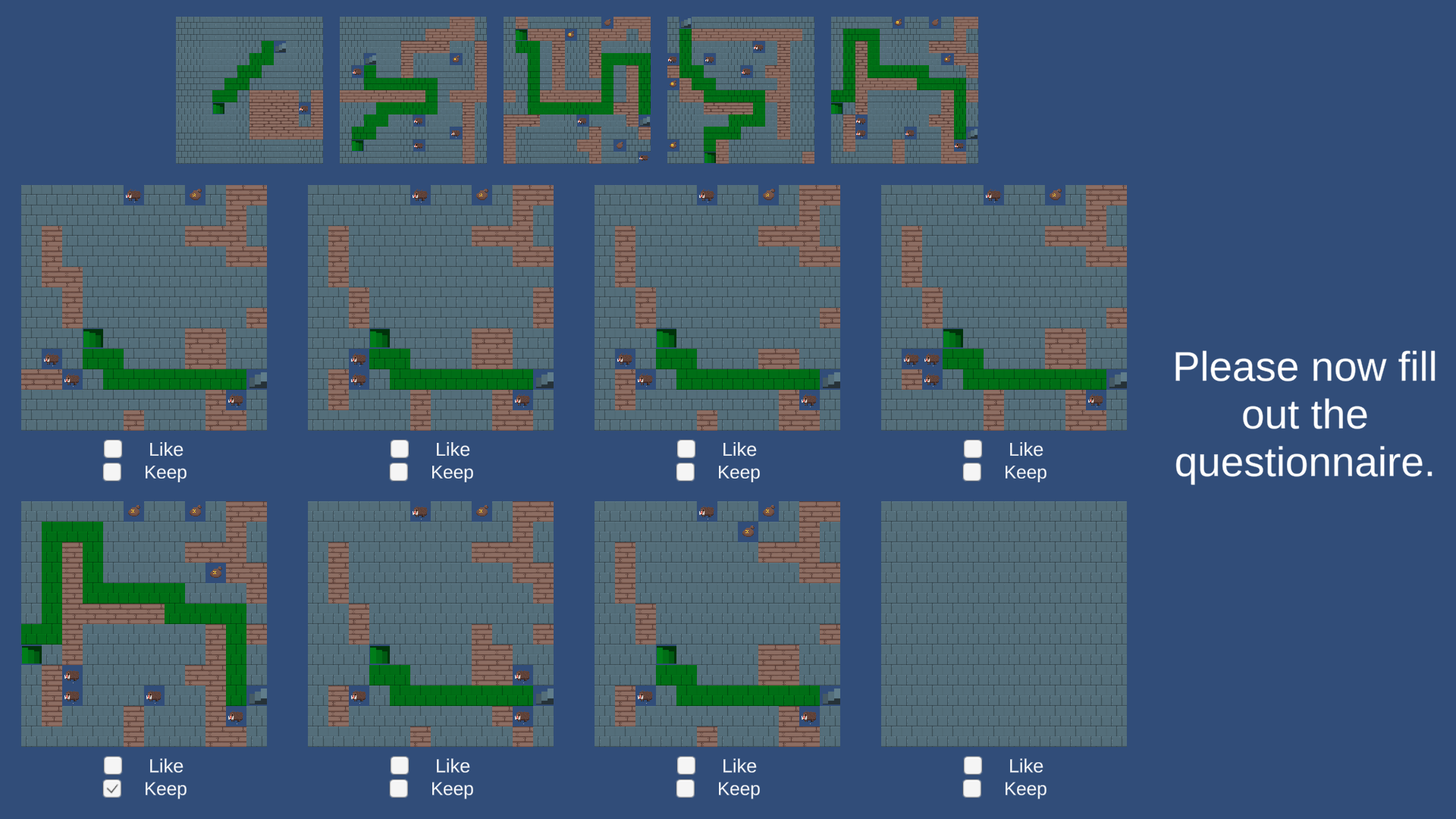}
    \caption{Genetic Algorithm Participant C}
    
\end{figure}
\begin{figure}[!t]
    \centering
    \includegraphics[width=3.5in]{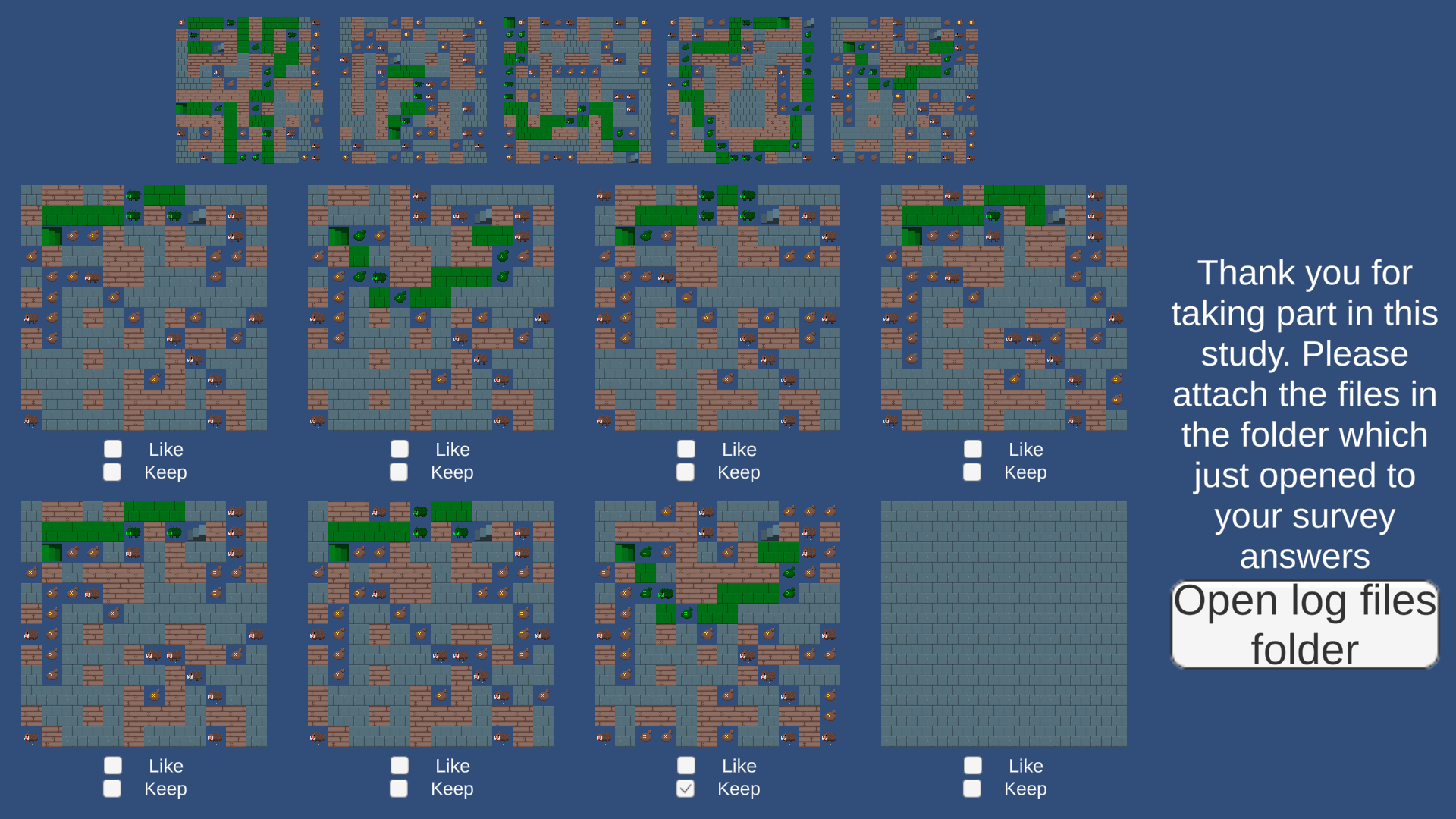}
    \caption{Genetic Algorithm Participant D}
    
\end{figure}
\begin{figure}[!t]
    \centering
    \includegraphics[width=3.5in]{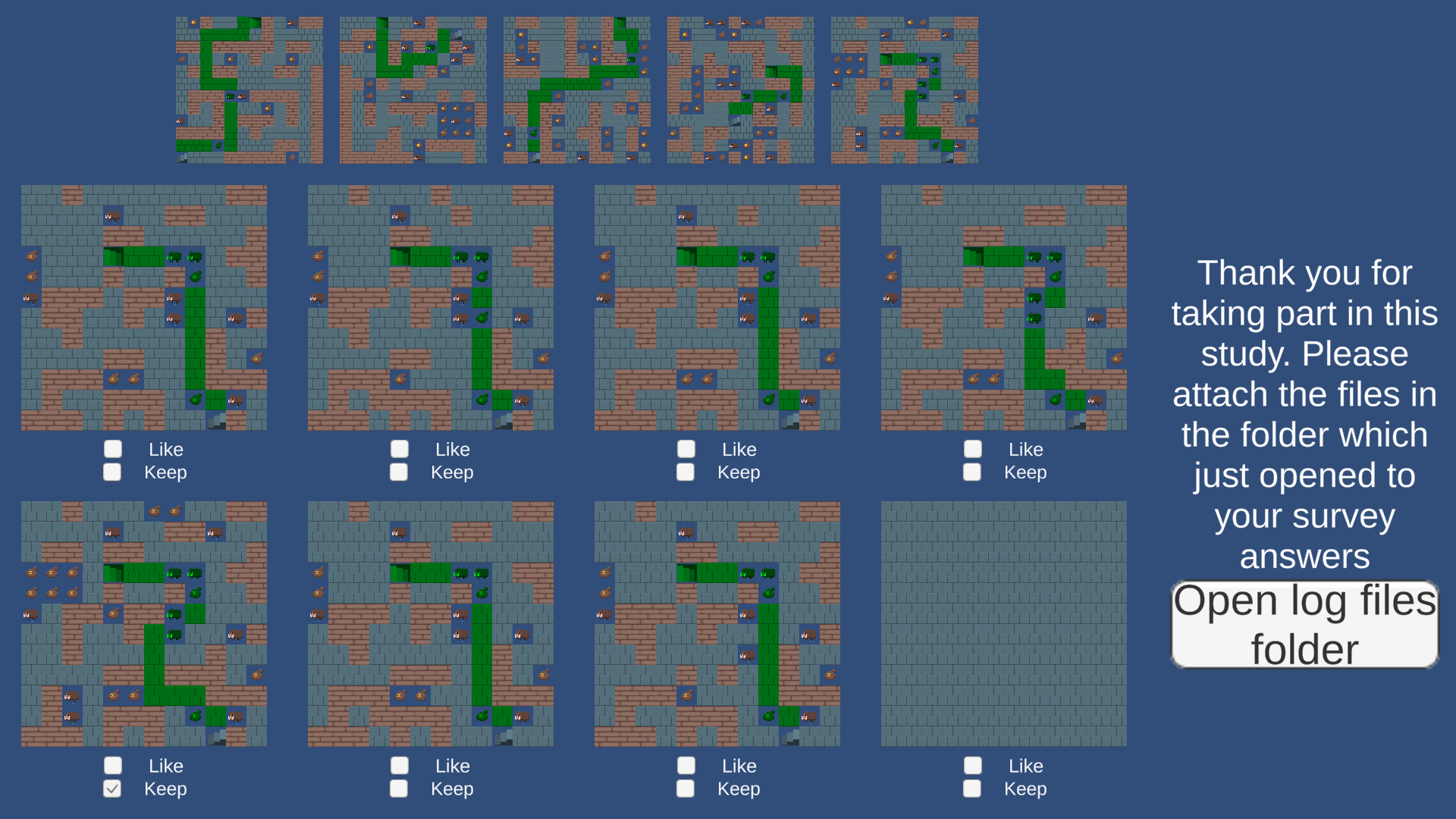}
    \caption{Genetic Algorithm Participant E}
    
\end{figure}
\begin{figure}[!t]
    \centering
    \includegraphics[width=3.5in]{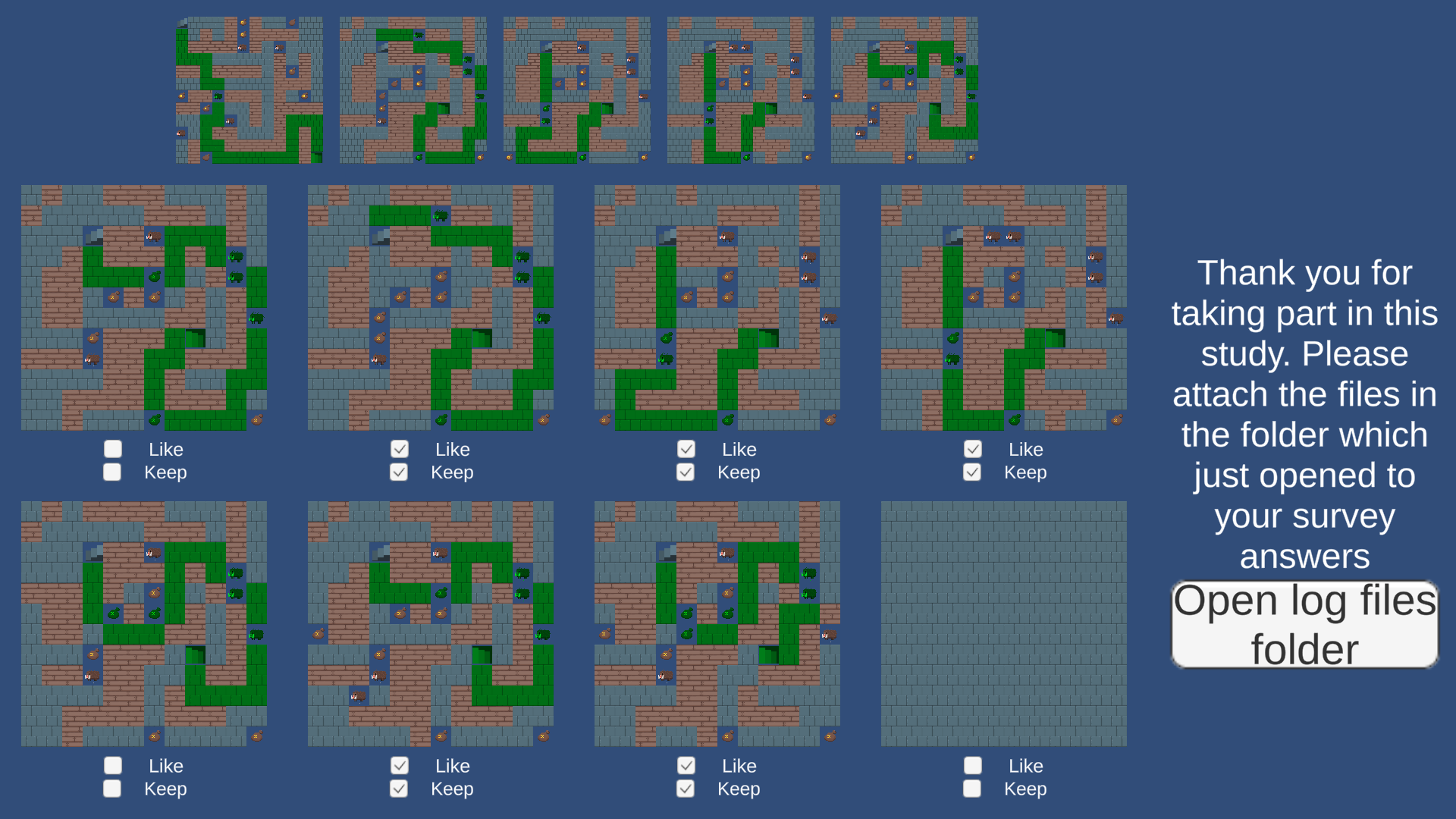}
    \caption{Genetic Algorithm Participant F}
    
\end{figure}
\begin{figure}[!t]
    \centering
    \includegraphics[width=3.5in]{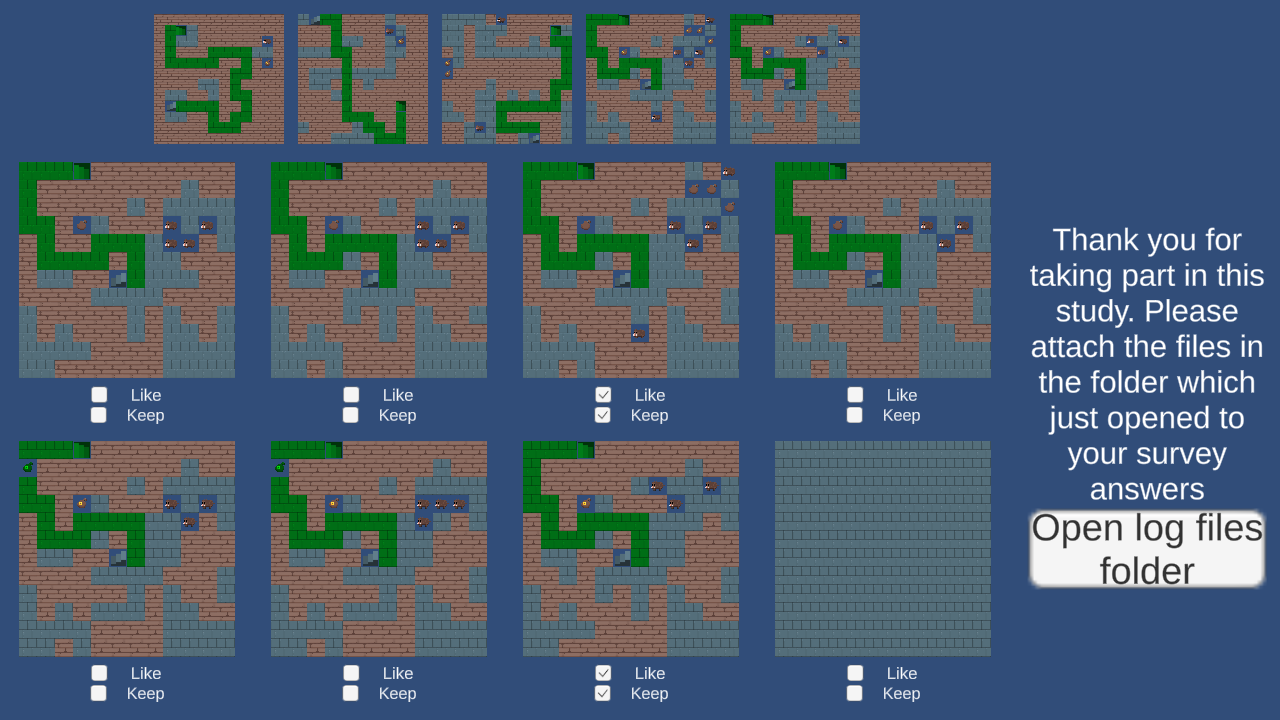}
    \caption{Genetic Algorithm Participant G}
    
\end{figure}
\begin{figure}[!t]
    \centering
    \includegraphics[width=3.5in]{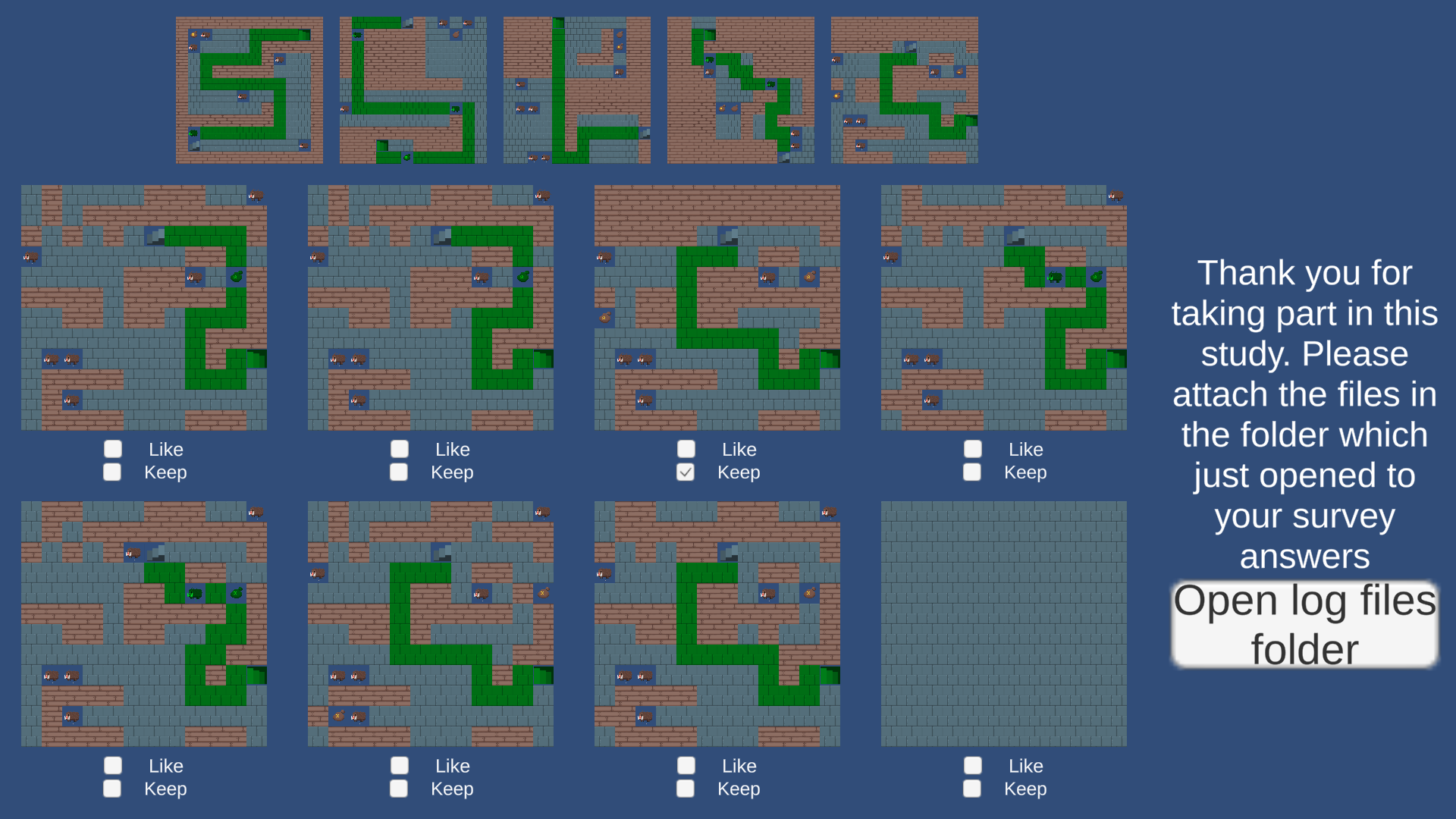}
    \caption{Genetic Algorithm Participant H}
    
\end{figure}
\begin{figure}[!t]
    \centering
    \includegraphics[width=3.5in]{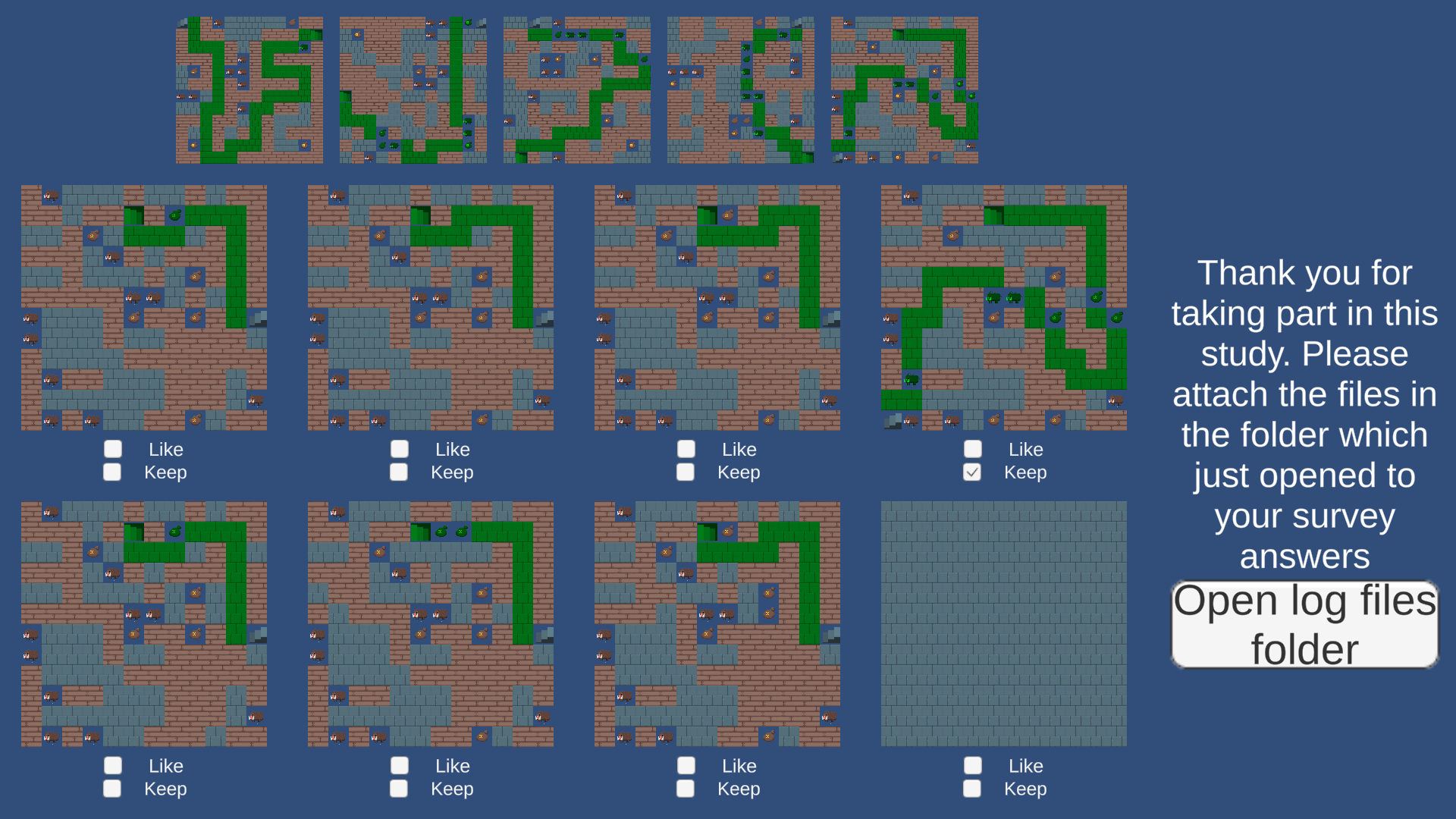}
    \caption{Genetic Algorithm Participant I}
    
\end{figure}
\begin{figure}[!t]
    \centering
    \includegraphics[width=3.5in]{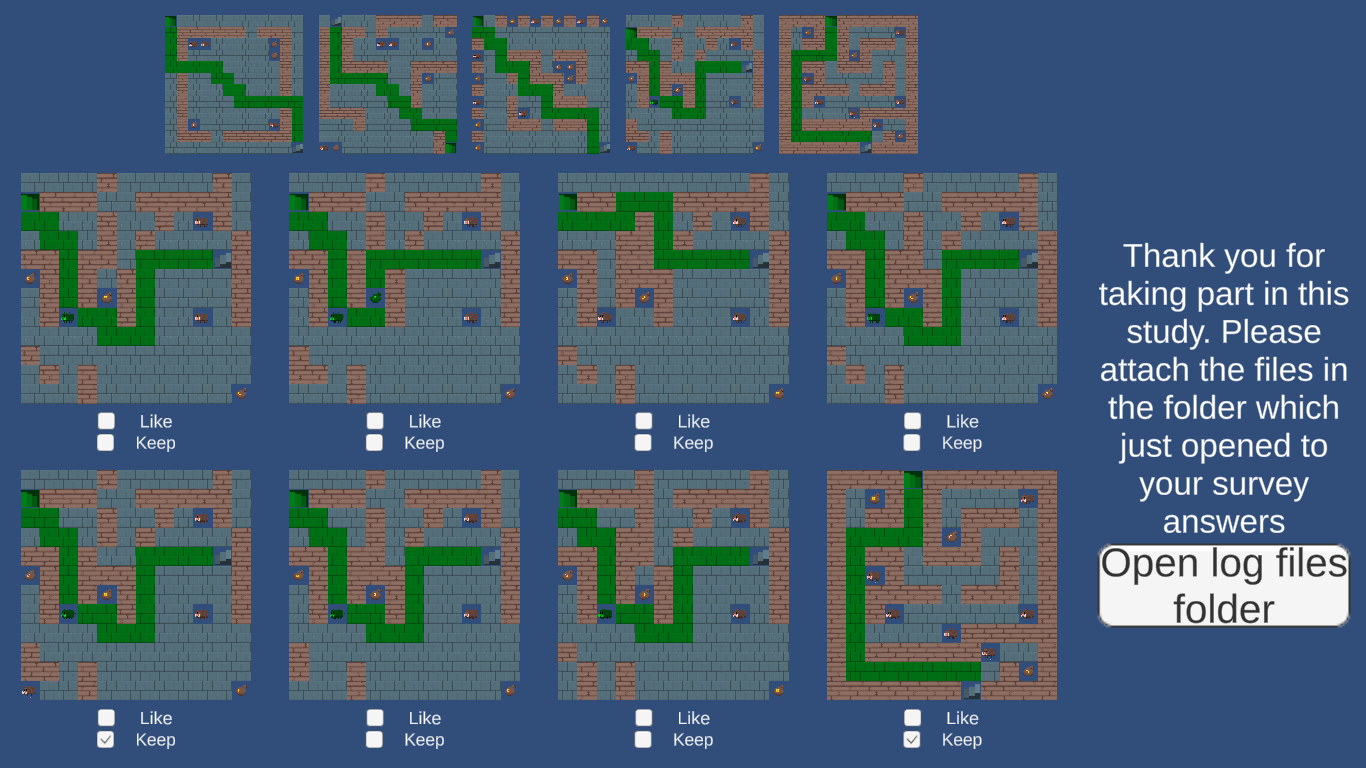}
    \caption{Genetic Algorithm Participant J}
    
\end{figure}
\begin{figure}[!t]
    \centering
    \includegraphics[width=3.5in]{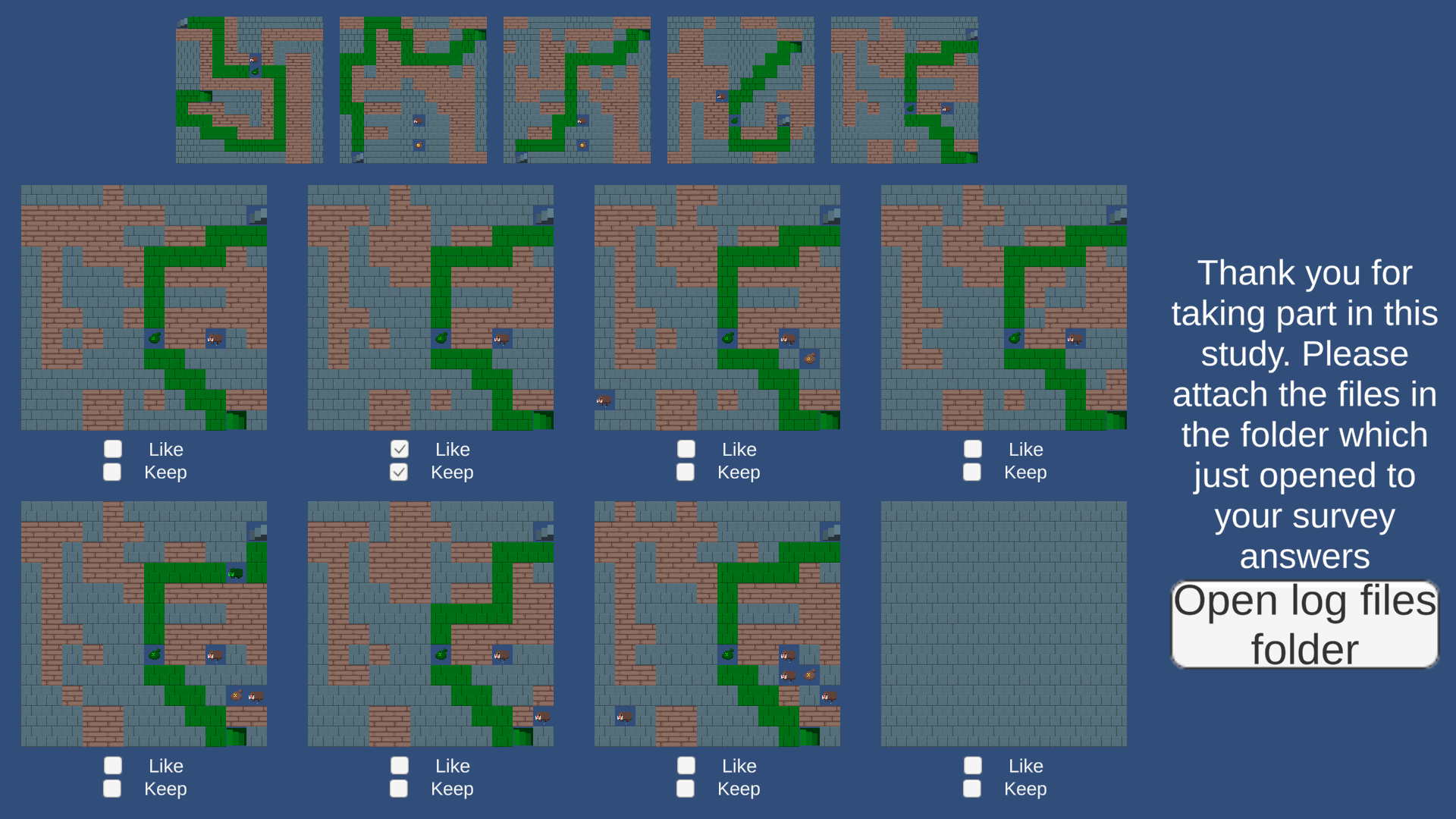}
    \caption{Genetic Algorithm Participant K}
    \label{GAK}
\end{figure}
\fi

\end{document}